# Stage 4 validation of the Satellite Image Automatic Mapper lightweight computer program for Earth observation Level 2 product generation – Part 2: Validation

A. Baraldi[a,c,]*, M. L. Humber[b], D. Tiede[c] and S. Lang[c]

[a] Department of Agricultural and Food Sciences, University of Naples Federico II, Portici (NA), Italy.
[b] Department of Geographical Sciences, University of Maryland, College Park, MD 20742, USA.
[c] Department of Geoinformatics – Z_GIS, University of Salzburg, Salzburg 5020, Austria.

*Corresponding author. Email: andrea6311@gmail.com

*Abstract* – The European Space Agency (ESA) defines an Earth Observation (EO) Level 2 product as a multi-spectral (MS) image corrected for geometric, atmospheric, adjacency and topographic effects, stacked with its data-derived scene classification map (SCM) whose legend includes quality layers such as cloud and cloud-shadow. No ESA EO Level 2 product has ever been systematically generated at the ground segment. To contribute toward filling an analytic and pragmatic information gap from EO big sensory data to the ESA EO Level 2 product, a Stage 4 validation (*Val*) of an off-the-shelf Satellite Image Automatic Mapper (SIAM) lightweight computer program for prior knowledge-based MS color naming was conducted by independent means. A time-series of annual Web-Enabled Landsat Data (WELD) image composites of the conterminous U.S. (CONUS) was selected as input dataset. The annual SIAM-WELD maps of the CONUS were validated in comparison with the U.S. National Land Cover Data (NLCD) 2006 map. These test and reference maps share the same spatial resolution and spatial extent, but their map legends are not the same and must be harmonized. For the sake of readability this paper is split into two. The previous Part 1 – Theory provided the multidisciplinary background of *a priori* color naming and proposed, first, an original guideline to identify a categorical variable-pair relationship based on a hybrid (combined deductive and inductive) inference approach; second, an original measure of categorical variable-pair association. The present Part 2 – Validation presents and discusses Stage 4 *Val* results collected from the test SIAM-WELD map time-series and the reference NLCD map by an original protocol for wall-to-wall thematic map quality assessment without sampling, where the test and reference map legends can differ in agreement with the Part 1. Conclusions are that the SIAM-WELD maps instantiate a Level 2 SCM product whose legend is the Food and Agriculture Organization of the United Nations – Land Cover Classification System (LCCS) taxonomy at the Dichotomous Phase (DP) Level 1 (vegetation/non-vegetation), Level 2 (terrestrial/aquatic) or superior LCCS level.

## 1. Introduction

For the sake of readability this paper is split into two. The preliminary Part 1 – Theory advanced the thesis that a necessary not sufficient pre-condition for a yet-unfulfilled Global Earth Observation

System of Systems (GEOSS) development, promoted by the intergovernmental Group on Earth Observations (GEO) (GEO 2005), is the systematic transformation at the ground segment of multi-source single-date multi-spectral (MS) imagery into a European Space Agency (ESA) EO Level 2 product (ESA 2015), instantiated as follows. (i) The MS image is corrected for geometric, atmospheric, adjacency and topographic effects (ESA 2015; CNES 2016). (ii) The enhanced MS image is stacked with its data-derived Scene Classification Map (SCM) (ESA 2015; CNES 2016). (iii) A general-purpose, user- and application-independent SCM's legend agrees with the 3-level 9-class "augmented" Dichotomous Phase (DP) taxonomy of the Food and Agriculture Organization of the United Nations (FAO) - Land Cover Classification System (LCCS) (Di Gregorio and Jansen 2000). Such an "augmented" land cover (LC) class taxonomy encompasses the canonical 3-level 8-class LCCS-DP legend in addition to a thematic layer "other" or "rest of the world" which includes, for example, quality layers such as cloud and cloud-shadow, see Figure 2-1. (iv) A GEO Stage 4 validation of the EO Level 2 outcome and process is considered mandatory to comply with the GEO Quality Assurance Framework for Earth Observation (QA4EO) calibration/validation (*Cal*/*Val*) requirements (GEO-CEOS 2010). By definition a GEO Stage 3 *Val* requires that "spatial and temporal consistency of the product with similar products are evaluated by independent means over multiple locations and time periods representing global conditions. In Stage 4 *Val*, results for Stage 3 are systematically updated when new product versions are released and as the time-series expands" (GEO-CEOS WGCV 2015).

No ESA EO Level 2 product has ever been systematically generated at the ground segment (ESA 2015; CNES 2016). To contribute toward filling an analytic and pragmatic information gap from multi-source EO big data to the ESA EO Level 2 product, the primary goal of this interdisciplinary study was to undertake an original (to the best of these authors' knowledge, the first) outcome and process Stage 4 *Val* of an expert system for top-down (deductive) MS reflectance space hyperpolyhedralization, specifically, an off-the-shelf Satellite Image Automatic Mapper™ (SIAM™) lightweight computer program. Implemented in operating mode in the C/C++ programming language, the SIAM application software is "lightweight" because it runs automatically (without human-machine interaction), in near real-time (it is non-iterative, either one-pass or two-pass, with a computational complexity increasing linearly with the image size) and in tile streaming mode (it requires a fixed runtime memory occupation) (Baraldi et al. 2006, 2010a, 2010b, 2010c, 2011, 2012a, 2012b, 2012c, 2012d, 2013, 2015, 2016; Baraldi and Humber 2015). In addition to running on laptop and desktop computers the SIAM lightweight computer program is eligible for use in a mobile software application. In addition to running on laptop and desktop computers the SIAM lightweight computer program is eligible for use in a mobile software application. Eventually provided with a mobile user interface, a mobile software application is a lightweight computer program specifically designed to run on web browsers and mobile devices, such as tablet computers and smartphones. The core of the non-iterative SIAM software pipeline is a one-pass prior knowledge-based decision tree (expert system) for MS reflectance space hyperpolyhedralization (quantization, partitioning) into static (non-adaptive-to-data) color names, see Figure 2-2. Presented in the remote sensing (RS) literature where enough information was provided for the implementation to be reproduced (Baraldi et al. 2006), the SIAM expert system for color naming is followed by a well-posed two-pass superpixel detector in the color map-domain (Dillencourt et al. 1992; Sonka et al. 1994) and a per-pixel vector quantization (VQ) error assessment for VQ quality assurance, in agreement with the QA4EO *Val* guidelines, refer to Figure 1-3 in the Part

1.
There is a long history of prior knowledge-based MS reflectance space partitioners for static color naming developed but never validated by space agencies, public organizations and private companies for use in hybrid EO-IUSs in operating mode. EO value-adding products and services targeted by existing hybrid EO-IUSs conditioned by static color naming encompass a large variety of low-level EO image enhancement tasks (Ackerman et al. 1998; Luo et al. 2008; Lück and van Niekerk 2016; Richter and D. Schläpfer 2012a; Richter and D. Schläpfer 2012b; Baraldi, Humber and Boschetti 2013; Baraldi and Humber 2015; Dorigo et al. 2009; Vermote and Saleous 2007; DLR and VEGA 2011; Lück and van Niekerk 2016; Baraldi et al. 2010c; Despini et al. 2014) and high-level EO image understanding applications (Baraldi et al. 2015; DLR and VEGA 2011; Lück and van Niekerk 2016; Muirhead and Malkawi 1989; Simonetti et al. 2015; GeoTerraImage 2015; Arvor et al. 2016; Boschetti et al. 2015; Baraldi et al. 2010a, 2010b). The potential impact on existing or future hybrid EO-IUSs in operating mode of an original (to the best of these authors' knowledge, the first) outcome and process Stage 4 *Val* of an off-the-shelf SIAM lightweight computer program for prior knowledge-based MS reflectance space hyperpolyhedralization, superpixel detection and per-pixel VQ quality assessment is expected to be relevant.

To comply with the GEO Stage 4 *Val* requirements an off-the-shelf SIAM application had to be validated by independent means on a radiometrically calibrated EO image time-series at large spatial extent. The open-access U.S. Geological Survey (USGS) 30 m resolution Web Enabled Landsat Data (WELD) annual composites of the conterminous U.S. (CONUS) for the years 2006 to 2009, radiometrically *Cal* into top-of-atmosphere reflectance (TOARF) values (Roy et al. 2010; Homer et al. 2004; WELD 2016) was identified as a viable input dataset. The 30 m resolution 16-class U.S. National Land Cover Data (NLCD) 2006 map, delivered in 2011 by the U.S. Geological Survey (USGS) Earth Resources Observation Systems (EROS) Data Center (EDC) (Vogelmann et al. 1998, 2001; Wickham et al. 2010; Wickham et al. 2013; Xian and Homer 2010; EPA 2007), was selected as the reference thematic map at the CONUS spatial extent. The 16-class NLCD map legend is described in Table 2-1. To account for typical non-stationary geospatial statistics, the NLCD 2006 thematic map was partitioned into 86 Level III ecoregions of North America collected from the Environmental Protection Agency (EPA) (EPA 2013; Griffith and Omernik 2009).

In the proposed experimental framework the test SIAM-WELD color map time-series and the reference NLCD 2006 map share the same spatial extent and spatial resolution, but their map legends are not the same. These working hypotheses are neither trivial nor conventional in the RS literature, where thematic map quality assessments typically adopt a sampling strategy, either random or non-random, and assume that the test and reference thematic map dictionaries coincide (Stehman and Czaplewski 1998). Starting from a stratified random sampling protocol presented in literature (Baraldi et al. 2013), the present Part 2 – Validation proposes an original protocol for wall-to-wall comparison without sampling of two thematic maps featuring the same spatial extent and spatial resolution, but whose legends can differ. This novel protocol incorporates two original contributions of the Part 1 where, first, a hybrid (combined deductive and inductive) guideline was proposed to streamline a human decision maker in the identification of a binary relationship R: A $\Rightarrow$ B between two univariate categorical variables A and B of the same population. This is an inherently ill-posed (equivocal, subjective) *information-as-data-interpretation* process (Capurro and Hjørland 2003) belonging to the multi-disciplinary domain of cognitive science, refer to Figure 1-12 in the Part 1. Second, version 2 of a categorical variable-pair association index (CVPAI2) $\in$ [0, 1] was proposed to cope with the entity-relationship conceptual model shown in Figure 1-16 of the Part 1.

The rest of the present Part 2 is organized as follows. Section 2 describes materials including the SIAM computer program, the time-series of annual WELD composites, the reference NLCD 2006 map and the EPA Level III ecoregion map of North America. Methods, specifically, an original protocol to compare without sampling the test SIAM-WELD and reference NLCD 2006 maps of the CONUS, whose map legends do not coincide and must be harmonized (reconciled, associated, translated (Ahlqvist 2005)), is proposed in Section 3. Experimental results are presented in Section 4 and discussed in Section 5. Conclusions are reported in Section 6.

## 2. Materials

Presented in the RS literature, four alternative implementations of a prior knowledge-based decision tree for static MS reflectance space hyperpolyhedralization onto static color names were compared for model selection. (i) The year 2006 SIAM decision tree presented in Baraldi et al. (2006). (ii) The static decision tree for Spectral Classification of surface reflectance signatures (SPECL), proposed by Dorigo et al. (2009), see Table 1-4 in the Part 1, and implemented by the Atmospheric/Topographic Correction for Satellite Imagery (ATCOR) commercial software product (Richter and D. Schläpfer 2012a, 2012b). (iii) The static decision tree for Single-Date Classification (SDC), proposed by Simonetti et al. (2015). (iv) The Canada Centre for Remote Sensing (CCRS) spectral decision tree shown in Figure 1-15 of the Part 1 (Luo et al. 2008). Whereas the SDC, SPECL and CCRS decision trees declare their applicability to Landsat images exclusively, SIAM claims its scalability to MS imaging sensors featuring different spectral resolution specifications, see Table 1-1 in the Part 1. Moreover, the SIAM decision tree outperforms its counterparts in terms of spectral quantization capability, parameterized by the total number of detected color names, equal to 96 for the 7-band Landsat-like SIAM (L-SIAM) subsystem, see Table 1-1 in the Part 1, versus 13, 19 and 7 color names detected in Landsat images by the SDC, SPECL (see Table 1-4 in the Part 1) and CCRS (see Figure 1-15 in the Part 1) decision trees respectively. To explain their broad differences in terms of number of detected color names and scalability to MS imaging sensors whose spectral and spatial resolution specifications can vary, the four static spectral decision trees of interest were compared at the level of understanding of spectral information/knowledge representation (Marr 1982), irrespective of the implementation of the decision rule set (structural knowledge in the decision tree) and of the order of presentation of decision rules (procedural knowledge in the decision tree).

To investigate the scalability of an *a priori* knowledge-based spectral decision tree to varying MS imaging sensor specifications we started observing that, given a partition of a MS color space $\Re^{MS}$ into a discrete and finite dictionary of hyperpolyhedra equivalent to color names {1, ColorDictionaryCardinality}, for any spatial unit x, either (0D) pixel, (1D) line or (2D) polygon defined according to the Open Geospatial Consortium (OGC) nomenclature (OGC 2015) and featuring a numeric ColorValue(x) $\in \Re^{MS}$, the photometric attribute of spatial unit x can be assigned with a categorical ColorName* $\in$ {1, ColorDictionaryCardinality}, such that membership m(ColorValue(x)| ColorName*) = 1, see Equation (1-3) in the Part 1. In practice any prior knowledge-based spectral decision tree for color naming can work at the sensor spatial resolution whatever it is, i.e., it can work pixel-based irrespective of the spatial resolution of the imaging sensor.

Since they are independent of the spatial resolution of the imaging sensor, static decision trees for color naming depend on spectral resolution specifications exclusively. Inter-sensor differences in spectral resolution can vary from minor differences in a band-specific sensitivity curve to the major lack of a whole spectral channel. To gain robustness to changes in spectral resolution specifications, the necessary not sufficient pre-condition for spectral rules is to infer "strong" (robust and reliable) conjectures based on the redundant convergence of multiple independent sources of spectral evidence, each of which is individually "weak". This rationale is alternative to, for example, pruning of redundant processing elements in distributed processing systems such as multi-layer perceptrons (Bishop 1995; Cherkassky and Mulier 1994). If this diagnosis holds true, i.e., redundancy of spectral evidence is a value-added of spectral rules to scale to varying spectral resolutions, then information redundancy of a spectral rule is expected to increase monotonically with the number of independent spectral terms.

In a MS reflectance space, any target family of LC class-specific spectral signatures is a multivariate data distribution (envelope, hyperpolyhedron, manifold). Like a vector quantity has two characteristics, a magnitude and a direction, any LC class-specific MS manifold is characterized by a multivariate shape and a multivariate intensity, see Figure 2-2. Hence, spectral information redundancy required to gain robustness to changes in spectral resolution specifications can regard the modelling of both the MS shape and MS intensity information components of a target MS manifold. Among the spectral decision trees being compared, only the SIAM decision tree adopts two different sets of spectral rules to model the MS shape and the MS intensity as two independent spectral information components of a target manifold of MS signatures. On the contrary, in the SDC, SPECL and CCRS decision trees MS shape and MS intensity properties are modeled simultaneously. For example, a typical SDC spectral rule applied to a Landsat pixel vector, radiometrically calibrated into a TOARF value in range [0, 1] in each Landsat band 1 to 6, is

If NDVI < 0.5 and NIR (= Landsat band 4) ≥ 0.15 then do something else do otherwise.

In this spectral decision rule, the normalized difference vegetation index, NDVI = (NIR – Red) / NIR + Red), where NIR = Landsat band 4 and Red = Landsat band 3, is a well-known spectral index, whose unbounded version is the band ratio NIR/Red. For example, band ratios are scalar spectral indexes widely employed in the SPECL decision tree, see Table 1-4 in the Part 1, and in the CCRS decision tree shown in Figure 1-15 of the Part 1 (Luo et al. 2008). Any scalar spectral index, either normalized band difference or band ratio, is conceptually equivalent to the slope of a tangent to the spectral signature in one point. This spectral slope is a MS shape descriptor independent of the MS intensity, i.e., infinite functions with different intensity values can feature the same tangent value in one point. Although appealing due to its conceptual and numerical simplicity (Liang 2004), any scalar spectral index is unable *per se* to represent either the multivariate shape information or the multivariate intensity information component of a MS signature. Intuitively a scalar spectral index causes a dramatic N-to-1 loss in spectral resolution by reducing an N-channel MS image to a panchromatic image. No photointerpreter whose objective is a single LC class detection, e.g., vegetation detection, would typically consider a panchromatic image as informative as a MS image or simple enough to be considered a binary image, e.g., vegetation/non-vegetation, where a simple crisp intensity thresholding criterion can be applied for target detection. In practice no univariate or multivariate spectral index is representative of the multivariate shape and multivariate intensity information components of a MS manifold, see Figure 2-2. This obvious but not trivial observation explains why, in spectral pattern recognition applications, lossy scalar spectral indexes are ever-increasing in number and variety in the endless search for yet-another scalar spectral index, supposedly more informative (Liang 2004; Baraldi et al. 2010a, 2010b). In the SDC rule reported above, the first spectral term, NDVI < 0.5, constrains a MS shape attribute; it is employed in logical combination with a second spectral term, where a MS intensity value is constrained as NIR ≥ 0.15. The conclusion is that, unlike SIAM's, neither the SDC nor the SPECL nor the CCRS decision tree decompose a target MS manifold into its multivariate shape and multivariate intensity information components to make each information component easier to be investigated by multivariate data analysis. In each of its two independent sets of spectral rules for MS shape and MS intensity modelling SIAM pursues redundancy of spectral terms as a value-added for spectral scalability. Possible combinations of these two independent sets of spectral rules make the SIAM decision tree implementations, starting from that proposed in pseudo-code in (Baraldi et al. 2006), capable of representing the multivariate shape and multivariate intensity information components of a target MS hyperpolyhedron, neither necessarily convex nor connected, as a converging combination of many independent $j$th-variable functions, with $j \in \{1$, total number $N$ of spectral channels$\}$. Multivariate data statistics are known to be more informative than a sequence of univariate data statistics. For example, maximum likelihood data classification, accounting for multivariate data correlation and variance (covariance), is typically more accurate than parallelepiped data classification whose rectangular decision regions, equivalent to a concatenation of univariate data constraints, poorly fit multivariate data in the presence of bivariate cross-correlation (Lillesand and Kiefer 1979). In the RS common practice, thanks to its spectral redundancy of multivariate data statistics the "master" 7-band Landsat-like SIAM (L-SIAM) decision tree can be down-scaled to cope with "slave" MS imaging sensors whose spectral resolution is inferior to, but overlaps with, Landsat's, see Table 1-1 in the Part 1 (Baraldi et al. 2010a, 2010b).

Based on this decision tree model comparison these authors concluded that the SIAM's peculiar design for multivariate spectral information representation and the SIAM implementation complexity, superior in terms of number of rules and number of terms per rule, appeared sufficient to justify the SIAM claims of a finer spectral quantization capability and a superior spectral scalability. Based on these conclusions an off-the-shelf SIAM application software was selected and considered worth of a Stage 4 *Val* to comply with the QA4EO *Cal/Val* requirements, refer to Section 1.

To pursue a Stage 4 *Val* of the SIAM application the 30 m resolution U.S. NLCD 2006 map was selected as reference LC map. When this experimental work was conducted the NLCD 2006 map was the most recent release of the U.S. NLCD map series developed by the USGS EDC (Vogelmann et al. 1998, 2001; Wickham et al. 2010; Wickham et al. 2013; Xian and Homer 2010; EPA 2007; Homer et al. 2015), see Figure 2-3. By now the U.S. NLCD map series comprises the NLCD 1992, 2001, 2001 Version 2.0, 2006 (released in 2011) and 2011 (released in 2015) editions. The timeliness from image collection to NLCD product delivery, which includes information layers such as tree cover fraction and impervious fraction, has steadily decreased from the about 5 years of the initial NLCD product. Made available for public access in a provisional version in Feb. 2011, the NLCD 2006 map was "based primarily on the unsupervised classification" of Landsat-5 Thematic Mapper (TM) and "Landsat-7 Enhanced Thematic Mapper (ETM)+ images acquired in circa 2006" (Xian and Homer 2010). It is a 30 m resolution raster product in the Albers Equal Area projection, which is the reference standard for continental scale cartography produced by U.S. agencies. Its legend consists of 16 LC classes defined according to the Level II LC classification system, refer to Table 2-1 (EPA 2007). Validation of the NLCD 2006 map provided an overall accuracy (OA) of 78%, which increased to 84% when the 16 LC classes were aggregated into 9 LC classes (Wickham et al. 2010; Wickham et al. 2013; Stehman et al. 2008). Noteworthy, these 9 LC classes are conceptually equivalent to an "augmented" 9-class LCCS-DP taxonomy, refer to Section 1. The validated NLCD 2006 map's OA values of 78% and 84% with, respectively, a 16 and a 9 LC class legend can be considered state-of-the-art. For example, they are superior to OAs featured by national-scale maps recently generated by pixel-based random forest classifiers from monthly WELD composites, whose OA is 65%–67% using 22 detailed classes and 72%–74% using 12 aggregated national classes (Wessels et al. 2016). In general renowned experts in Geographical Information Science (GIScience) suggest that "the widely used target accuracy of 85% may often be inappropriate and that the approach to accuracy assessment adopted commonly in RS can be pessimistically biased" (Foody 2006, 2016).

Based on these observations we considered the NLCD 2006 map's official OA estimate of 84% realistic and state-of-the-art at the U.S. national scale when the 3-level 9-class "augmented" LCCS-DP legend is adopted. As a consequence the reference NLCD 2006 map was considered suitable for a Stage 4 *Val* of the SIAM application software whose Level 2 SCM product's legend had to comply with the "augmented" 9-class LCCS-DP taxonomy. When a test SIAM map and a reference NLCD map share the same 30 m spatial resolution and spatial extent, then they can be compared wall-to-wall without sampling. Since no conventional sampling-theory procedure is employed (Lunetta and Elvidge 1999), a wall-to-wall OA(Test SIAM $\Rightarrow$ Reference NLCD) estimate in range [0, 100%] is provided with a confidence interval (degree of uncertainty in measurement), $\pm \delta \in$ [0, 100%], whose estimation is considered mandatory by the QA4EO *Val* guidelines, equal to $\pm \delta = 0\%$.

From a statistic standpoint the aforementioned experimental work specifications imply the following. Let's identify with OA(Test SIAM $\Rightarrow$ "Ultimate" GroundTruth) $\in$ [0, 100%] = 100% - Mismatch(Test SIAM $\Rightarrow$ "Ultimate" GroundTruth) the OA of an EO data-derived SIAM test map with respect to an "ultimate" (ideal) ground truth and with OA(Test SIAM $\Rightarrow$ Reference NLCD) $\pm$ 0% = 100% - Mismatch(Test SIAM $\Rightarrow$ Reference NLCD) $\pm$ 0% the overall degree of agreement provided with its confidence interval of a test SIAM map compared wall-to-wall without sampling with a reference NLCD map at the same spatial resolution and spatial extent. It is known that

OA(Reference NLCD 2006, "augmented" LCCS-DP 9 classes $\Rightarrow$ "Ultimate" GroundTruth 2006, "augmented" LCCS-DP 9 classes) = 84% = 100% - Mismatch(Reference NLCD 2006, "augmented" LCCS-DP 9 classes $\Rightarrow$ "Ultimate" GroundTruth 2006, "augmented" LCCS-DP 9 classes) = 100% - 16%.

Similarly,

OA(Reference NLCD 2006, NLCD 16 classes $\Rightarrow$ "Ultimate" GroundTruth 2006, NLCD 16 classes) = 78% = 100% - Mismatch(Reference NLCD 2006, NLCD 16 classes $\Rightarrow$ "Ultimate" GroundTruth 2006, NLCD 16 classes) = 100% - 22%.

Based on the superposition principle it is possible to write

OA(Test SIAM ⇒ "Ultimate" GroundTruth) ∈ [0, 100%] = OA(Test SIAM ⇒ Reference NLCD ⇒ "Ultimate" GroundTruth) = {OA(Reference NLCD ⇒ "Ultimate" GroundTruth) ± Mismatch(Test SIAM ⇒ Reference NLCD) ± 0%} ∈ [Worst Case, Best Case], where Worst Case = max{0%, Lower Bound} and Best Case = min{100%, Upper Bound}, with Lower Bound ≤ Upper Bound ∈ [0%, 100%],

where

Lower Bound = [OA(Reference NLCD 2006, "augmented" LCCS-DP 9 classes ⇒ "Ultimate" GroundTruth 2006, "augmented" LCCS-DP 9 classes) – Mismatch(Test SIAM ⇒ Reference NLCD 2006, "augmented" LCCS-DP 9 classes) ± 0%] =

[100% – Mismatch(Reference NLCD 2006, "augmented" LCCS-DP 9 classes ⇒ "Ultimate" GroundTruth 2006, "augmented" LCCS-DP 9 classes) – (100% - OA(Test SIAM ⇒ Reference NLCD 2006, "augmented" LCCS-DP 9 classes) ± 0%)] =

[OA(Test SIAM ⇒ Reference NLCD 2006, "augmented" LCCS-DP 9 classes) ± 0% - Mismatch(Reference NLCD 2006, "augmented" LCCS-DP 9 classes ⇒ "Ultimate" GroundTruth 2006, "augmented" LCCS-DP 9 classes)] =

[OA(Test SIAM ⇒ Reference NLCD 2006, "augmented" LCCS-DP 9 classes) ± 0% - 16%],

and

Upper Bound = [OA(Reference NLCD 2006, "augmented" LCCS-DP 9 classes ⇒ "Ultimate" GroundTruth 2006, "augmented" LCCS-DP 9 classes) + Mismatch(Test SIAM ⇒ Reference NLCD 2006, "augmented" LCCS-DP 9 classes) ± 0%] =

[84% + (100% - OA(Test SIAM ⇒ Reference NLCD 2006, "augmented" LCCS-DP 9 classes) ± 0%)] = [184% - OA(Test SIAM ⇒ Reference NLCD 2006, "augmented" LCCS-DP 9 classes) ± 0%].

To recapitulate, when the "Ultimate" GroundTruth adopts an "augmented" 9-class LCCS-DP legend, it is expected that

OA(Test SIAM ⇒ "Ultimate" GroundTruth 2006, "augmented" LCCS-DP 9 classes) ∈ [max{0%, Lower Bound}, min{100%, Upper Bound}] =

[max{0%, OA(Test SIAM ⇒ Reference NLCD 2006, "augmented" LCCS-DP 9 classes) ± 0% - 16%}, min{100%, 184% - OA(Test SIAM ⇒ Reference NLCD 2006, "augmented" LCCS-DP 9 classes) ± 0%}].     (2-1)

Similarly, when the "Ultimate" GroundTruth adopts a 16-class NLCD legend, then it is expected that

OA(Test SIAM ⇒ "Ultimate" GroundTruth 2006, NLCD 16 classes) ∈ [max{0%, Lower Bound}, min{100%, Upper Bound}] =

[max{0%, OA(Test SIAM ⇒ Reference NLCD 2006, NLCD 16 classes) ± 0% - 22%}, min{100%, 178% - OA(Test SIAM ⇒ Reference NLCD 2006, NLCD 16 classes) ± 0%}].     (2-2)

Equation (2-1) and Equation (2-2) are useful because, first, they highlight the undisputable fact that *per se* the NLCD reference map is not a "ground truth" for the SIAM test map, but only a reference baseline for comparison purposes. Second, they support the validity of this experimental project by showing that a summary statistic OA(Test SIAM ⇒ "Ultimate" GroundTruth) = OA(Test SIAM ⇒ Reference NLCD ⇒ "Ultimate" GroundTruth) can be inferred from an estimated OA(Test SIAM ⇒

Reference NLCD 2006) ± 0% known that OA(Reference NLCD 2006 ⇒ "Ultimate" GroundTruth 2006) is equal to 84% or 78% when the "Ultimate" GroundTruth adopts an "augmented" 9-class LCCS-DP legend or the 16-class NLCD legend respectively.

Supported by NASA and distributed by the USGS EDC (WELD 2016), the annual WELD composites for years 2006, 2007, 2008 and 2009 were selected as a large-scale radiometrically calibrated EO image time-series required by a Stage 4 *Val* of the SIAM application software in comparison with the reference NLCD 2006 map, see Figure 2-3. Each annual WELD composite consists of approximately 8,000 Landsat-5/7 image acquisitions per year over the CONUS, starting from year 2003 to 2012. The current WELD processing workflow requires as input Landsat sensor series L1T images with cloud cover ≤ 20%. The WELD mosaic of the CONUS encompasses 501 fixed location tiles defined in the Albers Equal Area projection. Each tile is 5000 × 5000 pixels in size, equal to 150 × 150 km (Homer et al. 2004). The Landsat sensor series L1T image geolocation error in the CONUS, including areas with substantial terrain relief, is less than 30 m (< 1 pixel) (Lee et al. 2004). The most recent Landsat data radiometric *Cal* expertise is employed in the WELD workflow to ensure harmonization and interoperability of multi-sensor Landsat image time-series, with a 5% absolute reflective band *Cal* uncertainty (Markham and Helder 2012), in agreement with the QA4EO's *Cal/Val* requirements (GEO-CEOS 2010). Figure 2-4 shows the WELD 2006 composite over the CONUS, where TOARF values are depicted in true colors, with the WELD tiling scheme overlaid in white.

To account for typical non-stationary geospatial statistics, an inter-map statistical comparison on a stratified (masked) basis should be accomplished at a local spatial extent, where strata convey some geospatial criteria of land surface information invariance. The NLCD 2006 reference map was partitioned into Level III ecoregions of North America collected from the EPA (EPA 2013). There are 86 ecoregions across the CONUS, each ecoregion featuring similar ecological and climatic characteristics (Griffith and Omernik 2009). Distributed as vector data, the EPA Level III ecoregions were rasterized to 30 m resolution in the Albers Equal Area projection. Figure 2-3 shows the NLCD 2006 map with boundaries of ecoregions overlaid in black.

## 3. Methods

A wall-to-wall comparison without sampling between the test SIAM-WELD map time-series and the reference NLCD 2006 map, sharing the same 30 m spatial resolution at the CONUS spatial extent, but whose legends A = DictionaryOfColorNames (see Table 2-2) and B = LegendOfObjectClassNames (see Table 2-1) do not coincide and must be harmonized, was designed and implemented for Stage 4 *Val* purposes. These working hypotheses differ from thematic map accuracy assessment protocols adopted by the large majority of the RS community, typically based on an either random or non-random sampling and/or a confusion matrix (CMTRX). A CMTRX is defined as a special case of a two-way contingency table (bivariate table), BIVRTAB = FrequencyCount(A × B) where A × B is a 2-fold Cartesian product between two univariate categorical variables A and B estimated from the same population (Kuzera and Pontius 2008; Pontius and Connors 2006). In particular a CMTRX is square and sorted because the test and reference categorical variables A and B of the same population are required to be the same, to let the main diagonal guide the interpretation process, refer to the Part 1, Section 2.

In (Baraldi et al. 2014), a crisp thematic map assessment protocol was proposed based on probability sampling, a pair of test and reference thematic legends A and B that may differ, an overlapping area matrix (OAMTRX) (Beauchemin and Thomson 1997; Ortiz and Oliver 2006), whose spatial unit x is (0D) pixel, a set of thematic quantitative quality indicators ($Q^2$Is), T$Q^2$Is, extracted from the OAMTRX and a set of spatial $Q^2$Is (S$Q^2$Is) extracted from sub-symbolic image-objects in the multi-level map domain, where image-objects are either (0D) pixels, (1D) lines or (2D) polygons according to the OGC nomenclature (OGC 2015). Equivalent to an either square or non-square BIVRTAB = FrequencyCount(A × B), an OAMTRX is generated from the cross-tabulation of any possible pair of univariate categorical variables A and B, either coincident or not, estimated from the same geospatial population, refer to Part 1, Section 2 (Beauchemin and Thomson 1997; Ortiz and Oliver 2006). Whereas the construction of an OAMTRX is straightforward and non-controversial when the semantic labels of sampling units are crisp (hard), the method to construct an OAMTRX when semantic labels are soft (fuzzy) is not obvious at all, e.g., refer to (Kuzera and Pontius 2008). Hence, we focused on crisp OAMTRX instances, exclusively. To accomplish our work hypotheses, the crisp thematic map probability sampling protocol proposed in (Baraldi et al. 2014) was modified as follows.

o The original hybrid eight-step guideline proposed in the Part 1, Section 4 was adopted to streamline the inherently subjective selection by human experts of a binary relationship R: A = DictionaryOfColorNames $\Rightarrow$ B = LegendOfObjectClassNames $\subseteq$ A × B that guides the interpretation processs of a crisp OAMTRX = FrequencyCount(A × B), see Table 1-3 in the Part 1.

o Given a binary relationship R: A = DictionaryOfColorNames $\Rightarrow$ B = LegendOfObjectClassNames that guides the interpretation process of a crisp OAMTRX = FrequencyCount(A × B), a novel CVPAI2(R: A $\Rightarrow$ B) formulation was adopted as a relaxed version of the CVPAI1 formulation proposed in (Baraldi et al. 2014), refer to the Part 1, Section 5.

o Traditional 30 m resolution Landsat image classifiers are pixel-based, due to the lack of contextual information in 30 m resolution imagery. Hence, in the 30 m resolution WELD composites, the most informative planar entity is (0D) pixel, rather than image-object, either (1D) line or (2D) polygon (OGC 2015). As a consequence, for the sake of simplicity, in the present thematic map comparison image-object-based S$Q^2$Is were omitted. Rather, the following pixel-based T$Q^2$Is were estimated from the crisp OAMTRX = FrequencyCount(A × B) estimated wall-to-wall with spatial unit x equal to pixel.

- An OA(OAMTRX) ± 0% was computed in line with (Baraldi et al. 2014). This OA estimate is guided by the binary relationship R: A = DictionaryOfColorNames $\Rightarrow$ B = LegendOfObjectClassNames identified and community-agreed upon in advance, refer to this text above. In an OAMTRX estimated from a wall-to-wall inter-map comparison, where no sample data is investigated, any adopted T$Q^2$I features a degree of uncertainty in measurement equal to ± 0%, e.g., see Equation (2-1).

- User's and producer's accuracies, computed in (Baraldi et al. 2014), were replaced by class-conditional probabilities, *p(r | t)* of reference class *r* given test class *t* and, vice versa, *p(t | r)* of test class *t* given reference class *r*, with *r = 1, ..., RC*, and *t = 1, ..., TC*, where *RC* = |B| = b =

ObjectClassLegendCardinality and *TC* = |A| = a = ColorDictionaryCardinality are the total numbers of reference and test classes respectively,

The proposed ensemble of summary statistics, specifically, CVPAI2(R: A $\Rightarrow$ B $\subseteq$ A × B), OA(OAMTRX = FrequencyCount(A × B)) and class-conditional probabilities(OAMTRX), is an original minimally dependent and maximally informative (mDMI) set (Si Liu et al. 2011; Peng et al. 2005) of TQ$^2$Is, to be jointly maximized according to the Pareto formal analysis of multi-objective optimization problems (Boschetti et al. 2004), refer to the Part 1, Section 1.

## 4. Validation session

*Val* is the process of assessing (GEO-CEOS WGCV 2015), by independent means, the quality of an information processing system by means of an mDMI set (Si Liu et al. 2011; Peng et al. 2005) of community-agreed outcome and process (OP) Q$^2$Is provided with a degree of uncertainty in measurement, ±δ, in compliance with the QA4EO *Cal/Val* guidelines (GEO-CEOS 2010). The SIAM-WELD data mapping process and outcome were validated by a human expert independent of the present authors (refer to Acknowledgments). This independent human expert accomplished the following tasks. (I) Run without user interaction an off-the-shelf SIAM application upon the 30 m resolution annual WELD 2006 to 2009 composites of the CONUS. (II) Overlap the test SIAM-WELD annual map time-series with the reference NLCD 2006 map to generate instances of an OAMTRX = FrequencyCount(A × B) (Baraldi et al. 2014). (III) Estimate an mDMI set of OP-Q$^2$Is, encompassing the following, refer to the Part 1, Section 1 (Baraldi and Humber 2015; Baraldi et al. 2013; Baraldi and Boschetti 2012a, 2012b). (i) Product effectiveness. Proposed outcome Q$^2$Is (O-Q$^2$Is) were the TQ$^2$Is presented in Section 3: (a) CVPAI2(R: A $\Rightarrow$ B $\subseteq$ A × B; b) OA(OAMTRX = FrequencyCount(A × B)), and (c) class-conditional probabilities $p(r | t)$ and p($t | r$) with reference class $r = 1$, ..., *RC* = |B| = ObjectClassLegendCardinality and test class $t = 1$, ..., *TC* = |A| = ColorDictionaryCardinality. (ii) Process efficiency. Proposed process Q$^2$Is (P-Q$^2$Is) were computation time and memory occupation. (iii) Process degree of automation, monotonically decreasing with the number of system's free-parameters to be user-defined. (iv) Process robustness to changes in the input dataset. For post-classification change/no-change detection (Lunetta and Elvidge 1999), the SIAM-WELD 2006 to 2009 maps were compared one another when one year apart. (v) Process robustness to changes in input parameters, if any. (vi) Process scalability, to keep up with changes in users' needs and sensor properties. (vii) Product timeliness, defined as the time between data acquisition and product generation. (viii) Product costs in manpower and computer power. In the present study the following definition is adopted: an information processing system can be considered in operating mode (ready-to-go) if it scores "high" in all of its OP-Q$^2$I values, refer to the Part 1, Section 1.

For the sake of paper simplicity, the following decisions were undertaken.

o The SIAM-WELD 2006 color maps at fine (96) color names and intermediate (48) color names were compared with the NLCD 2006 map, while the SIAM map at coarse (18) color names was ignored, see Table 1-1 in the Part 1. This implied the following.
- In the case of the SIAM-WELD 2006 map at fine discretization level, an OAMTRX instance consisted of test set A = 96 spectral categories as rows and reference set B = 16 NLCD classes as columns. Due to its excessive size, this OAMTRX instance cannot be shown in a technical paper. However, it is made available on an anonymous ftp site (SIAM-WELD-NLCD FTP 2016) and its TQ$^2$I summary statistics are reported in the present paper.
- In the case of the SIAM-WELD 2006 map at the intermediate color discretization level, an OAMTRX instance consisted of test set A = 48 spectral categories as rows and reference set B = 16 NLCD classes as columns. Due to its excessive size, this OAMTRX instance cannot be shown in a technical paper. Hence, the ensemble of 48 spectral categories were reassembled into 19 spectral macro-categories by the independent human expert, refer to Table 2-2. This grouping of basic color names into spectral macro-categories of basic colors pertains to the inherently equivocal (subjective) domain of *information-as-data-interpretation*, refer to the Part 1, Section 4 (Capurro and Hjørland 2003). Among the 19 spectral macro-categories reassembled by the independent human expert, 16 macro-categories coincided exactly with one category in the initial set of 48 spectral categories. The reassembled macro-category "*Others*" included the SIAM intermediate spectral category "*Unknowns*" together with 24 of the 48 intermediate spectral categories covering "disturbances", such as cloud, smoke plume, fire front, etc., which are typically minimized or removed in an annual WELD composite. The reduced set of 19 spectral macro-categories is mutually exclusive and totally exhaustive, in line with the Congalton and Green requirements of a classification scheme (Congalton and Green 1999). A simplified OAMTRX instance of reduced size was generated. It consisted of a test set A = 19 spectral macro-categories as rows and a reference set B = 16 NLCD classes as columns. Due to its reduced size, this OAMTRX instance can be shown in the present paper, together with its TQ$^2$I estimates.
o In agreement with the previous paragraph, the annual SIAM-WELD 2006 to 2009 maps at the SIAM intermediate discretization level of 48 color names were all reassembled into 19 spectral macro-categories, see Table 2-2.

### *4.1 Verification of the Co-Registration Requirements for Pixel-based Inter-Map Comparison*

In the requirements specification of RS projects dealing with per-pixel post-classification change/no-change detection, the required RS image co-registration error is typically < 1 pixel. For example, in (Lunetta and Elvidge 1999), it is recommended that the root-mean-square (RMS) error between any two-date images should not exceed 0.5 pixels.

In (Dai and Khorram 1998), simulated misregistration effects are investigated upon multi-temporal Landsat images of North Carolina across four study areas representative of land cover types: forest land, agricultural land, bare soil and urban/residential area. In these experiments, a registration accuracy < 1/5 of a pixel is considered necessary to achieve a land cover change detection error < 10%. This conclusion is more severe than the one-pixel co-registration constraint typically adopted in most change detection applications.

The annual WELD composites and the NLCD 2006 thematic map were derived from the same sensory dataset of Landsat L1T images acquired by the USGS EDC. It means that the SIAM-WELD 2006 pre-classification maps and the NLCD 2006 reference map were derived from the same sensory dataset. Hence, it is reasonable to assume that the co-registration error between these data-derived maps is negligible.

*4.2 Inter-Annual SIAM-WELD Map Comparisons for Years 2006 to 2009*

The consistency across time and space of the annual SIAM-WELD 2006 to 2009 map time-series with a legend of 19 spectral macro-categories was investigated. Based on *a priori* knowledge of the multi-temporal pixel-based selection criteria adopted by the USGS EDC for the generation of annual WELD composites (refer to Section 2) and of the LC/LC change (LCC) dynamics in the real-world CONUS, a small percentage of LCC counts was expected to be detected one year apart at the CONUS spatial extent.

Table 2-3 shows percentages of the CONUS assigned to each SIAM-WELD spectral macro-category across the annual time-series. The green-as-"*Vegetation*" spectral categories are predominant (refer to the total vegetation statistic reported in Table 2-3), with an average 79% of the CONUS pixels, followed by color names such as brown-as-"*Bare soils or built-up*" (19% on average), followed by the remaining spectral macro-categories which, altogether, account for about 2%. The standard deviation through time of the occurrence of each SIAM-WELD spectral macro-category at the CONUS spatial extent is lower than 1%, with the exception of two vegetation spectral categories (specifically, aV_HC and aV_MC) where a larger variance can be attributed mostly to phenology. If a vegetation-through-time spectral variability due to changes in phenology affects the annual WELD composites then the data-derived SIAM-WELD color quantization maps will be affected by changes in phenology too. This diagnosis was verified as follows. Due to the limited availability of cloud-free Landsat observations at a generic pixel location per year, the Julian day of the year of the observation selected at a given location (pixel) of the WELD composite changes through years (Roy et al. 2010). This is illustrated in Figure 2-5 where, at any fixed location across a target "ground-truth" area of deciduous forest in a pair of monthly August-November WELD composites, the SIAM spectral labels change significantly, but consistently with the phenological season. The same consideration holds when changes in phenology affect the annual WELD composites. This can explain the "high" intra-vegetation spectral variability observed by the SIAM vegetation macro-categories aV_HC and aV_MC in the tested time-series of annual WELD composites.

Non-stationary spatial phenomena occurring at the CONUS spatial extent can be oversighted by global statistics. To be captured spatial non-stationarities require local statistics such as class-conditional statistics described in Table 2-3. According to Section 3, for every pair of one annual SIAM-WELD test map with legend A = 19 spectral macro-categories for year 2006 to 2009 overlapped at the CONUS spatial extent with a reference NLCD 2006 map with legend B = NLCD 16 classes, the pair of summary statistics CVPAI2(R: A $\Rightarrow$ B $\subseteq$ A $\times$ B) $\in$ [0, 1.0] and OA(OAMTRX = FrequencyCount(A $\times$ B)) $\in$ [0, 1.0] should be maximized jointly. Shown as gray entry-pair cells in Table 2-4, the binary relationship R: A $\Rightarrow$ B selected by the independent human expert featured a CVPAI2(R: A $\Rightarrow$ B) = 0.6769 while the OA(OAMTRX) = OA(Test SIAM-WELD 2006, 19 spectral macro-categories $\Rightarrow$ Reference NLCD 2006, NLCD 16 classes) = 96.88% ± 0%. With a binary relationship R: A $\Rightarrow$ B kept fixed, where CVPAI2(R: A $\Rightarrow$ B) = 0.6769, the OA(OAMTRX) estimate became equal to 97.02%, 96.69% and 96.75% for the annual SIAM-WELD map of year 2007 to 2009 compared with the reference NLCD 2006 map.

*4.3 Comparison of the SIAM-WELD 2006 and NLCD 2006 Thematic Maps*

The SIAM-WELD 2006 test maps at intermediate (48) and fine (96) color quantization levels were compared with the NLCD 2006 reference map as described below.

**Test case A**. The SIAM-WELD 2006 map of the CONUS at the intermediate color discretization level reassembled into 19 spectral macro-categories is shown in Figure 2-6. The OAMTRX instance generated from the overlap between the test SIAM-WELD 2006 map with legend A = 19 spectral macro-categories at the CONUS spatial extent with a reference NLCD 2006 map with legend B = NLCD 16 classes is shown in Table 2-4, where each cell reports a joint probability value $p$(SIAM-WELD$_t$, NLCD$_r$), $r$ = 1, ..., $RC$ = |B| = 16, refer to Table 2-2, and $t$ = 1, ..., $TC$ = |A| = 19, refer to Table 2-1. Gray entry-pair cells identify the binary relationship R: A $\Rightarrow$ B $\subseteq$ A $\times$ B chosen by the independent human expert to guide the OAMTRX interpretation process. The distribution of these "correct" entry-pairs shows that every NLCD class overlaps with several discrete color types, with the exceptions of two SIAM-NLCD entry-pairs, specifically, entry-pair [SIAM spectral macro-category, NLCD class] = [MS white-as-"*Snow*" (SN, see Table 2-2), NLCD class "*Perennial ice/snow*" (PIS, see Table 2-1)] and entry-pair [SIAM spectral macro-category, NLCD class] = [MS blue-as-"*Water or Shadow*" (WA, see Table 2-2), NLCD class "*Open water*" (OW, see Table 2-1)], which are both characterized by a 1-1 matching relation. According to their specific definitions (refer to Table 2-1), anthropic NLCD classes, such as "*Developed, Open Space*" (DOS), "*Developed, Low Intensity*" (DLI), "*Developed, Medium Intensity*" (DMI) and "*Developed, High intensity*" (DHI), are a mixture of vegetated surfaces, impervious surfaces and bare soil, in agreement with the popular vegetation-impervious surface-soil model for urban ecosystem analysis (Ridd 1995). In agreement with their definitions, these NLCD classes overlap exclusively with the SIAM spectral macro-categories related to vegetation or bare soil. The NLCD class "*Barren Land*" (BL, see Table 2-1) overlaps with all of the SIAM spectral macro-categories related to bare soil. Noteworthy, according to Table 2-4, the NLCD class BL covers only 1.21% of the CONUS total surface. This is due to the NLCD 2006 definition of class BL (Rock/Sand/Clay), very restrictive with regard to the presence of vegetation, which has to account for less than 15% of total cover. The NLCD definition of class BL means that the NLCD classes "*Shrub/Scrub*" (SS) and "*Grassland/Herbaceous*" (GH, refer to Table 2-1) may feature a vegetated cover which accounts for 15% of total cover or more. The NLCD forest classes "*Deciduous forest*" (DF), "*Evergreen Forest*" (EF) and "*Mixed forest*" (MF, refer to Table 2-1) overlap with the SIAM's high and medium canopy-cover spectral macro-categories. The NLCD vegetation classes "*Shrub/Scrub*" (SS) and "*Grassland/Herbaceous*" (GH, refer to Table 2-1) overlap with the SIAM-WELD 2006 medium and low canopy-cover spectral macro-categories, but, in case of dry or sparse vegetation, also with some of the SIAM-WELD 2006 spectral macro-categories related to bare soil, namely, sbS_1, SmS_1 and aS (refer to Table 2-2). The overlap between the reference NLCD 2006 vegetation classes SS and GH and the test SIAM-WELD 2006 bare soil spectral macro-categories sbS_1, SmS_1 and aS is the only case of comprehensive (systematic) "semantic mismatch" recorded across the wall-to-wall SIAM-WELD 2006 and NLCD 2006 thematic map comparison. Hence, it is worth a deeper analysis in comparison with an "ultimate" ground truth. Reported in this section above, the NLCD 2006 definition of class "*Barren Land*" (BL, see Table 2-1) means that classes SS and GH may feature a vegetated cover which accounts for 15% of total cover or more. Two consequence of these definitions are that, whereas the NLCD class BL covers only 1.21% of the CONUS total surface, the NLCD vegetation classes SS and GH map the near totality of desert areas across the CONUS. Hence, there is a systematic "semantic mismatch" between the NLCD 2006 vegetation classes SS and GH and the SIAM-WELD 2006 bare soil spectral macro-categories across nearly all desert areas of the CONUS. Figure 2-7 shows real-world examples of geographic locations mapped as vegetation classes "*Scrub/Shrub*" (SS) or "*Grassland/Herbaceous*" (GH) in the NLCD 2006 map (refer to Table 2-1), while they are mapped predominantly as the bare soil spectral categories sbS_1, SmS_1 and aS in the SIAM-WELD 2006 map (refer to Table 2-2). For more comments about this systematic case of "conceptual mismatch" between the test SIAM-WELD and reference NLCD 2006 maps, refer to Figure 2-10.

Additional inter-map overlaps highlighted by Table 2-4 reveal that the NLCD class "*Pasture/Hay*" (PH, see Table 2-1) occurs together with high and medium canopy-cover color types of the SIAM-WELD map. The NLCD class "*Cultivated crops*" (CC, see Table 2-1) matches with both spectral macro-categories MS green-as-"*Vegetation*" and MS brown-as-"*Bare soil or built-up*". Finally, the NLCD classes of wetland ("*Woody Wetlands*", WW, and "*Emergent Herbaceous Wetland*", EHW, see Table 2-1) overlap with the SIAM's vegetated spectral macro-categories or with spectral macro-category MS blue-as-"*Water or Shadow*" (WA, refer to Table 2-2).

As reported in Section 4.2, in the OAMTRX instance shown in Table 2-4 summary statistics are CVPAI2(R: A $\Rightarrow$ B) = 0.6769 and OA(OAMTRX) = OA(Test SIAM-WELD 2006, 19 spectral macro-categories $\Rightarrow$ Reference NLCD 2006, NLCD 16 classes) = 96.88% ± 0%. As a consequence, according to Equation (2-2),

> OA(Test SIAM 2006, 19 spectral macro-categories $\Rightarrow$ "Ultimate" GroundTruth 2006, NLCD 16 classes) $\in$ [max{0%, Lower Bound}, min{100%, Upper Bound}] =
> [max{0%, OA(Test SIAM 2006, 19 spectral macro-categories $\Rightarrow$ Reference NLCD 2006, NLCD 16 classes) ± 0% - 22%}, min{100%, 178% - OA(Test SIAM 2006, 19 spectral macro-categories $\Rightarrow$ Reference NLCD 2006, NLCD 16 classes) ± 0%}] =
> [max{0%, 96.88% ± 0% - 22%}, min{100%, 178% - 96.88% ± 0%}] =
> [74.88%, 81.12%], with CVPAI2(R: A = SIAM dictionary of 19 color names $\Rightarrow$ B = NLCD legend of 16 LC class names) = 0. 6769, hence the semantic information gap from sub-symbolic sensory data to symbolic NLCD classes left to be filled by further stages in the EO-IUS pipeline = 1 – CVPAI2 = 0.3231, refer to the Part 1, Section 5. (2-3)

When disagreements between the two reference and test maps were back-projected onto the WELD 2006 image domain, these specific WELD sites were photointerpreted by the independent human expert to provide an additional independent source of thematic evidence for Stage 4 *Val* of the SIAM-WELD 2006 test map. The large majority of the CONUS areas where the NLCD vegetation classes overlap with the SIAM spectral macro-category MS blue-as-"*Water or Shadow*" (WA, refer to Table 2-2) or, vice versa, where the SIAM vegetation spectral macro-categories overlap with the NLCD reference class "*Open Water*" (OW, refer to Table 2-1) were identified by the independent human photointerpreter as riparian zones. In practice, these riparian zones were labeled by the SIAM-WELD 2006 and NLCD 2006 maps in two different conditions of their annual surface status. Also in this case the SIAM-WELD labeling appears consistent with the human photointerpretation of the WELD composite, irrespective of the semantic disagreement between this SIAM-WELD labeling and the NLCD 2006 reference map.

Based on evidence collected by the independent photointerpreter with regard to the systematic "conceptual mismatches" between the test SIAM-WELD 2006 map and the reference NLCD 2006 map across nearly all desert areas and riparian zones of the CONUS, validated conclusions were twofold. First, according to Equation (2-1) where the NLCD 2006 reference map is not a "ground truth" for the SIAM-WELD 2006 test map, but only a reference baseline for comparison purposes, inter-map "conceptual mismatches" should not be misinterpreted as mapping errors by the test SIAM-WELD 2006 map with respect to an "ultimate" ground truth. Second, an OA(Test SIAM 2006, 19 spectral macro-categories $\Rightarrow$ "Ultimate" GroundTruth 2006, NLCD 16 classes) summary statistic, inferred to belong to range [74.88%, 81.12%] according to Equation (2-3), was considered likely to be lying closer to its upper bound.

**Test case B**. To reveal the inherent ill-posedness of any inter-dictionary "conceptual matching" (refer to the Part 1, Section 4), one co-author of this paper, different from the independent human expert (refer to Acknowledgments), conducted a second inherently equivocal selection of "correct" entry-pairs in the binary relationship R: A $\Rightarrow$ B which guides the interpretation process of the OAMTRX instance shown in Table 2-4. This second experiment provided a CVPAI2(R: A $\Rightarrow$ B) = 0.6804 and an OA(OAMTRX) = 97.28% $\pm$ 0%, which are both superior to (better than) the pair of summary statistics CVPAI2(R: A $\Rightarrow$ B) = 0.6769 and OA(OAMTRX) = 96.88% $\pm$ 0% provided by the independent human expert in the test case A. This binary relationship looks sparser and therefore less intuitive to understand than that shown as dark entry-pair cells in Table 2-4. Hence, it is not shown in this paper, although it is made available via anonymous ftp (SIAM-WELD-NLCD FTP 2016). When these summary statistics replace variables in Equation (2-2), we obtain

OA(Test SIAM 2006, 19 spectral macro-categories $\Rightarrow$ "Ultimate" GroundTruth 2006, NLCD 16 classes) $\in$ [max{0%, Lower Bound}, min{100%, Upper Bound}] =

[max{0%, OA(Test SIAM 2006, 19 spectral macro-categories $\Rightarrow$ Reference NLCD 2006, NLCD 16 classes) $\pm$ 0% - 22%}, min{100%, 178% - OA(Test SIAM 2006, 19 spectral macro-categories $\Rightarrow$ Reference NLCD 2006, NLCD 16 classes) $\pm$ 0%}] =

[max{0%, 97.28% $\pm$ 0% - 22%}, min{100%, 178% - 97.28% $\pm$ 0%}] =

[75.28%, 80.72%], with CVPAI2(R: A = SIAM dictionary of 19 color names $\Rightarrow$ B = NLCD legend of 16 LC class names) = 0.6804, hence the semantic information gap from sub-symbolic sensory data to symbolic NLCD classes left to be filled by further stages in the EO-IUS pipeline = 1 – CVPAI2 = 0.3196, refer to the Part 1, Section 5.   (2-4)

**Test case C**. The wall-to-wall overlap between the test SIAM-WELD 2006 map at fine (96) color discretization (refer to Table 1-1 in the Part 1) and the reference NLCD 2006 map generated another OAMTRX instance featuring a test set A = 96 color names as rows and a reference set B = 16 NLCD classes as columns, available via anonymous ftp (SIAM-WELD-NLCD FTP 2016). Again, the hybrid inference procedure described in the Part 1, Section 4 was employed by the independent human expert to select "correct" entry-pairs in the binary relationship R: A $\Rightarrow$ B that guides the interpretation process of this OAMTRX instance, whose estimated TQ$^2$I summary statistics became CVPAI2(R: A $\Rightarrow$ B) = 0.5809 and OA(OAMTRX) = 95.41% $\pm$ 0%. When these summary statistics replace variables in Equation (2-2), we obtained

OA(Test SIAM 2006, 96 spectral macro-categories ⇒ "Ultimate" GroundTruth 2006, NLCD 16 classes) ∈ [max{0%, Lower Bound}, min{100%, Upper Bound}] =

[max{0%, OA(Test SIAM 2006, 96 spectral macro-categories ⇒ Reference NLCD 2006, NLCD 16 classes) ± 0% - 22%}, min{100%, 178% - OA(Test SIAM 2006, 96 spectral macro-categories ⇒ Reference NLCD 2006, NLCD 16 classes) ± 0%}] =

[max{0%, 95.41% ± 0% - 22%}, min{100%, 178% - 95.41% ± 0%}] =

[73.41%, 82.59%], with CVPAI2(R: A = SIAM dictionary of 96 color names ⇒ B = NLCD legend of 16 LC class names) = 0.5809, hence the semantic information gap from sub-symbolic sensory data to symbolic NLCD classes left to be filled by further stages in the EO-IUS pipeline = 1 – CVPAI2 = 0.4191, refer to the Part 1, Section 5.   (2-5)

**Test case D**. When the NLCD classification taxonomy becomes less discriminative (coarser) because reassembled from 16 LC class names to 9 and 4 LC class names respectively, it is a fact that the following inequality holds.

OA(Reference NLCD 2006, NLCD 16 classes ⇒ "Ultimate" GroundTruth 2006, NLCD 16 classes) = 78% ≤

OA(Reference NLCD 2006, "augmented" LCCS-DP 9 classes ⇒ "Ultimate" GroundTruth 2006, "augmented" LCCS-DP 9 classes) = 84% ≤

OA(Reference NLCD 2006, 2-level LCCS-DP 4 classes ⇒ "Ultimate" GroundTruth 2006, 2-level LCCS-DP 4 classes) = XX%,   (2-6)

where the 2-level LCCS-DP 4 classes are (see Figure 2-1):

- A1 = Primarily Vegetated Terrestrial Areas = Cultivated Areas (A11) or (Semi) Natural Vegetation (A12).
- A2 = Primarily Vegetated Aquatic or Regularly Flooded Areas = Cultivated Aquatic Areas (A23) or (Semi) Natural Aquatic Vegetation (A24).
- B3 = Primarily Non-vegetated Terrestrial Areas = Artificial Surfaces (B35) or Bare Areas (B36).
- B4 = Primarily Non-vegetated Aquatic or Regularly Flooded Areas = Artificial (B47) or Natural Waterbodies, Snow and Ice (B48).

In agreement with Equation (2-1), Equation (2-2) and Equation (2-6) we could write

OA(Test SIAM 2006, 19 spectral macro-categories ⇒ "Ultimate" GroundTruth 2006, 2-level LCCS-DP 4 classes) ∈ [max{0%, Lower Bound}, min{100%, Upper Bound}] =

[max{0%, OA(Test SIAM 2006, 19 spectral macro-categories ⇒ Reference NLCD 2006, 2-level LCCS-DP 4 classes) ± 0% - (100% - XX%)}, min{100%, 100% + XX% - OA(Test SIAM 2006, 19 spectral macro-categories ⇒ Reference NLCD 2006, 2-level LCCS-DP 4 classes) ± 0%}], with XX% = OA(Reference NLCD 2006, 2-level LCCS-DP 4 classes ⇒ "Ultimate" GroundTruth 2006, 2-level LCCS-DP 4 classes) ≥ 84%.   (2-7)

As shown in the test case A, according to the NLCD definitions (refer to Table 2-1), LC classes "*Developed, Open Space*" (DOS), "*Developed, Low Intensity*" (DLI), "*Developed, Medium Intensity*" (DMI) and "*Developed, High intensity*" (DHI) are a spatial mixture of vegetated surfaces, impervious surfaces and bare soil, in agreement with the popular vegetation-impervious surface-soil model for urban ecosystem analysis (Ridd 1995). It means that a logical OR combination of the NLCD classes DOS or DLI or DMI or DHI mainly matches with the 2-level LCCS-DP classes B3 or A1. In general, it is not possible to backtrack (reconstruct) each of the 2-level LCCS-DP 4 classes starting from the NLCD 16 class taxonomy without ambiguity. To transform a reference NLCD map with an NLCD 16-class legend into a reference NLCD map with a 2-level LCCS-DP 4-class legend we adopted the approximated binary relationship R: A $\Rightarrow$ B $\subseteq$ A $\times$ B, with set A = NLCD 16-class legend and set B = 2-level LCCS-DP 4-class legend, reported in Table 2-5 and summarized below.

- A1 = Primarily Vegetated Terrestrial Areas = Cultivated Areas or (Semi) Natural Vegetation $\approx$ NLCD 16-classes DF (41) or EF (42) or MF (43) or SS (52) or GH (71) or PH (81) or CC (82). Actually, this NLCD class OR-combination is an expected mixture of LCCS-DP classes A1 and B3 as first- and second-best match respectively.
- A2 = Primarily Vegetated Aquatic or Regularly Flooded Areas = Cultivated Aquatic Areas or (Semi) Natural Aquatic Vegetation $\approx$ NLCD 16-classes WV (90) or EHW (95).
- B3 = Primarily Non-vegetated Terrestrial Areas = Artificial Surfaces or Bare Areas $\approx$ NLCD 16-classes DOS (21) or DLI (22) or DMI (23) or DHI (24) or BL (31). Actually, this NLCD class OR-combination is an expected mixture of LCCS-DP classes B3 and A1 as first- and second-best match respectively.
- B4 = Primarily Non-vegetated Aquatic or Regularly Flooded Areas = Artificial or Natural Waterbodies, Snow and Ice $\approx$ NLCD 16-classes OW (11) or PIS (12).

Fixed this approximated binary relationship R: A $\Rightarrow$ B $\subseteq$ A $\times$ B with set A = NLCD 16-class legend and set B = 2-level LCCS-DP 4-class legend, Table 2-5 shows the following.

- A binary relationship R: C $\Rightarrow$ B $\subseteq$ C $\times$ B with set C = SIAM 19-class legend as rows and set B = 2-level LCCS-DP 4-class legend as columns, where "correct" entry-pairs are shown as grey cells in the 2-fold Cartesian product C $\times$ B. This binary relationship was (subjectively) selected by the present authors according to the categorical variable-pair relationship identification strategy proposed in the Part 1, Section 5.
- An OAMTRX = FrequencyCount(C $\times$ B) generated by the wall-to-wall overlap between the test SIAM map with legend C and the reference NLCD map with legend B.

Table 2-5 provided a CVPAI2(R: C $\Rightarrow$ B) = 0.7486 and an OA(OAMTRX = FrequencyCount(C $\times$ B)) = 93.09% $\pm$ 0%. When these summary statistics replace variables in Equation (2-7), we obtained

OA(Test SIAM 2006, 19 spectral macro-categories $\Rightarrow$ "Ultimate" GroundTruth 2006, 2-level LCCS-DP 4 classes) $\in$ [max{0%, Lower Bound}, min{100%, Upper Bound}] =
[max{0%, 93.09% $\pm$ 0% - (100% - XX%)}, min{100%, 100% + XX% - 93.09% $\pm$ 0%}] =
[XX% - 6.91%, XX% + 6.91%], where XX% = OA(Reference NLCD 2006, 2-level LCCS-DP 4 classes $\Rightarrow$ "Ultimate" GroundTruth 2006, 2-level LCCS-DP 4 classes) $\geq$ 84%, with CVPAI2(R: C = SIAM dictionary of 19 spectral macro-categories $\Rightarrow$ B = 2-level LCCS-DP 4 classes) = 0.7486, hence the semantic information gap from sensory data to the 2-level LCCS-DP 4-class legend left to be filled by further stages in the EO-IUS pipeline = 1 – CVPAI2 = 0.2514, refer to the Part 1, Section 5.                    (2-8)

*4.4 Probabilities of the SIAM-WELD Test Labels Conditioned by the NLCD Reference Labels and Vice Versa*

Table 2-4 shows the OAMTRX instance generated from the wall-to-wall overlap of the SIAM-WELD 2006 test map featuring 19 spectral macro-categories, see Table 2-2 and Figure 2-6, with the NLCD 2006 reference map featuring 16 LC classes, see Table 2-1 and Figure 2-3. The division of each probability cell of Table 2-4 by its column-sum generates the conditional probability $p$(SIAM-WELD$_t$ / NLCD$_r$) of the SIAM-WELD 2006 test spectral category $t$, with $t = 1, ..., TC = 19$, given the NLCD 2006 reference class $r$, with $r = 1, ..., RC = 16$, refer to Figure 2-8 and Table 2-6, where Table 2-6 is a summarized text version of Figure 2-8. To prove their plausibility, conditional probabilities $p$(SIAM-WELD$_t$ / NLCD$_r$), $t = 1, ..., TC, r = 1, ..., RC,$ should agree with theoretical expectations stemming from human experience. For instance, it was expected that the NLCD 2006 reference classes "*Deciduous Forest*" (DF), "*Evergreen Forest*" (EF) and "*Mixed Forest*" (MF), refer to Table 2-1, overlap with vegetated spectral categories in the test SIAM-WELD 2006 map, while the NLCD reference class "*Developed, High Intensity*" (DHI, see Table 2-1) was expected to be mostly matched by bare soil macro-categories in the test SIAM-WELD 2006 map. Overall, these prior knowledge-based expectations about specific class-conditional probabilities appear satisfied by both Figure 2-8 and Table 2-6.

In the RS common practice, once a generic user has generated at no cost in manpower and computer power, i.e., in near real-time and without user-machine interaction, a SIAM color map from an unknown EO image, what this user wishes to do is to infer from the EO image a set of LC classes (say, "*Forest*"), conditioned by the detected SIAM spectral categories (say, MS green-as-"*Vegetation*"). To accomplish this spectral category-conditional inference, class-conditional probabilities $p$(SIAM-WELD$_t$ / NLCD$_r$), $t = 1, ..., TC, r = 1, ..., RC$, shown in Table 2-6, are not useful. Rather, this generic user can found helpful to know the conditional probabilities of an NLCD 2006 reference class $r$, with $r = 1, ..., RC = 16$, given the SIAM-WELD 2006 spectral category $t$, with $t = 1, ..., TC = 19$. These are the class-conditional probabilities $p$(NLCD$_r$ | SIAM-WELD$_t$), $t = 1, ..., TC, r = 1, ..., RC$, generated by dividing each probability cell of Table 2-4 by its row-sum. They are shown in Figure 2-9 and summarized in text form in Table 2-7. Very intuitive to understand, Table 2-7 clearly highlights the two main semantic inconsistencies found between the reference NLCD 2006 and test SIAM-WELD 2006 maps already reported in Section 4.3. First, the SIAM vegetation spectral macro-category wV_HC ("*Weak evidence vegetation with high canopy cover*", refer to Table 2-2) is best-matched by the reference NLDC class "*Open Water*" (OW, refer to Table 2-2). Since this semantic mismatch occurs almost exclusively in the CONUS areas recognized by the independent human expert as riparian zones typically depicted as mixed pixels at 30 m resolution, then the 30 m resolution SIAM labeling can be considered reasonable, if we consider that the crisp SIAM implementation is not expected to accomplish pixel unmixing. Second, the NLCD reference class "*Shrub/Scrub*" (SS, refer to Table 2-2) appears to be the best match for several of the SIAM bare soil spectral macro-categories. Figure 2-7 shows examples of geographic locations where this semantic mismatch occurs. In these locations, 30 m resolution pixels are typically affected by mixed spectral contributions that the crisp SIAM implementation is not expected to unmix.

*4.5 Stratification by Ecoregions*

Due to the central limit theorem, the arithmetic mean of a large number of independent random variables tends to be a Gaussian distribution, where independent "local" data distributions (like basis functions) become indistinguishable from the whole. For example, in human vision, the neural computations are inherently spatially local in the (2D) image-domain; next, a global spatial average is superimposed on the local computational processes. In general, non-stationary local features do not survive the averaging process, i.e., the precise position of each local contribution is no longer perceived after the averaging process (Victor 1994). Since the WELD composite of the CONUS is about ten billion pixels in size, summary statistics of the SIAM mapping quality at the CONUS spatial extent are inadequate to demonstrate the local-scale capability of the SIAM expert system to correctly map EO images, characterized by non-stationary local statistics. To investigate the SIAM mapping capability at local spatial extent, the SIAM-WELD 2006 and NLCD 2006 maps were stratified using the 86 EPA Level III ecoregions of the CONUS (see Figure 2-3) and an individual OAMTRX was generated per ecoregion. All 86 ecoregion-specific OAMTRX instances are available as supplemental online material (SIAM-WELD-NLCD FTP 2016). As one example of an inter-map comparison at the ecoregion spatial scale of analysis, let us consider the SIAM-WELD 2006 and NLCD 2006 maps of the Wyoming Basin ecoregion, which is predominantly desert, see Figure 2-10 where the ecoregion boundary is highlighted in red. Table 2-8 reports the corresponding OAMTRX instance. Table 2-8 shows that the predominantly desertic Wyoming Basin ecoregion is predominantly classified as the LC classes "*Scrub/Shrub*" (SS) and "*Grassland/Herbaceous*" (GH) in the NLCD 2006 reference map (refer to Table 2-1) and as bare soil spectral categories (sbS_1, SmS_1, aS) in the SIAM-WELD 2006 test map (refer to Table 2-2). This semantic disagreement was already observed in Section 4.3, also refer to Figure 2-7.

Figure 2-11 provides a synthetic representation of the full dataset of 86 ecoregion-specific OAMTRX instances (SIAM-WELD-NLCD FTP 2016). It shows for each of the 16 reference NLDC classes, with index $r = 1, ..., RC = 16$, the box-and-whisker diagram of the NLCD-class-conditional probabilities $p(\text{SIAM-WELD}_{er,t} / \text{NLCD}_{er,r})$, with $t = 1, ..., TC = 19$, collected across the 86 ecoregions, each ecoregion identified with an index $er = 1, …, ER = 86$. In each of the $TC = 19$ boxes of an NLCD class-specific boxplot, the median (shown as a horizontal line within the box) represents the general trend of the distribution and the dispersion around it describes the distribution variability across ecosystems. A small dispersion around the median value indicates a reference-to-test class mapping whose occurrence is nearly constant across ecosystems, while a large dispersion around the median indicates that occurrences of this inter-map relationship change significantly across ecosystems.

*4.5 OP-$Q^2I$ values of the SIAM application and product*

OP-$Q^2$Is of the SIAM application (refer to the introduction to Section 4) input with the 30 m resolution annual WELD 2006 to 2009 composites of the CONUS were collected by the independent human expert (refer to Acknowledgments). They are summarized below (Duke 2016).

(i) Process degree of automation. In line with theoretical expectations about expert systems (refer to Section 1), the SIAM application required neither user-defined parameters nor reference samples to run. Hence, its ease of use cannot be surpassed by any alternative inference approach.

(ii)     Outcome effectiveness. An mDMI set of O-$Q^2$Is (Si Liu et al. 2011; Peng et al. 2005), comprising a CVPAI2(R: A $\Rightarrow$ B), an OA(OAMTRX = FrequencyCount(A × B)), the class-conditional probabilities $p(r \mid t)$ of reference class $r = 1, ..., RC = |B|$, given test class $t = 1, ..., TC = |A|$, and class-conditional probabilities $p(t \mid r)$, with $r = 1, ..., RC = |B|, t = 1, ..., TC = |A|$, was estimated in the four test cases described in Section 4.3.

(iii)    Process efficiency: memory occupation and computation time. About memory occupation, the SIAM computer program adopts a tile streaming implementation, where the dynamic memory (random access memory, RAM) maximum occupation is a known function of the tile size to be fixed in advance, irrespective of the image size. In these experiments the RAM maximum occupation was set equal to 800 MB, which can be considered a "small" RAM value. About computation time: Run on a Dell Power Edge 710 server with dual Intel Xeon @ 2.70 GHz processor with 64 GB of RAM and a 64-bit Linux operating system, the SIAM application required less than 45 seconds to generate its complete set of per-image output products from a 7-band Landsat-7 ETM+ WELD tile of 5000 × 5000 pixels, which means about 8 hours to map an annual WELD composite of the CONUS. In our data mapping workflow, such an output rate was not inferior to the input rate of an annual WELD composite being implemented or delivered to end-users. Hence, the SIAM computation time was considered equivalent to near real-time, where the SIAM computational complexity increases linearly with the image size.

(iv)    Process robustness to changes in the input dataset. The SIAM mapping consistency of the annual WELD composites from year 2006 to 2009 was estimated to be "high" at the CONUS spatial extent, refer to Section 4.2 to Section 4.5.

(v)     Process robustness to changes in input parameters, if any. Since SIAM requires no user-defined parameter to run, its robustness to changes in input parameters cannot be surpassed by alternative approaches.

(vi)    Process maintainability/ scalability/ re-usability, to keep up with changes in users' needs and sensor properties. The multi-source SIAM physical model can be applied to any existing or future planned spaceborne/airborne MS imaging sensor provided with a radiometric calibration metadata file, refer to the existing literature (Baraldi and Humber 2015; Baraldi et al. 2010c; Baraldi et al. 2010a, 2010b) and to the Part 1, Section 2.

(vii)   Outcome timeliness, defined as the time span between data acquisition and product generation. Since it is prior knowledge-based and near real-time, the SIAM application reduces timeliness from image acquisition to color map generation to almost zero.

(viii)  Outcome costs, monotonically increasing with manpower and computer power. Since it is prior knowledge-based and near real-time in a standard laptop computer, the SIAM costs are almost negligible.

## 5. Discussion

Table 2-3 shows that the 30 m resolution annual SIAM-WELD map time-series at the CONUS spatial extent with an intermediate color discretization legend of 48 color names reassembled into 19 spectral macro-categories features a standard deviation of the annual frequency counts collected for each spectral macro-category lower than 1%, with the exception of two vegetated spectral macro-categories, specifically, aV_HC and aV_MC (see Table 2-2). These two larger spectral category-specific variations in annual frequency counts at the CONUS spatial extent can be attributed mostly to vegetation phenology. This was proved in Section 4.2: changes in phenology affect the monthly WELD and annual WELD composites and, as a consequence, the data-derived SIAM-WELD maps. These numerical results agree with the *a priori* knowledge of RS experts about the CONUS surface dynamics, whose inter-annual LCC summary statistics are expected to score low. The conclusion is that observations stemming from the annual SIAM-WELD map time-series with a legend of 19 spectral macro-categories comply with the domain knowledge of RS experts about the LC and LCC dynamics in the physical CONUS.

The interpretation process of the OAMTRX = FrequencyCount(A × B) shown in Table 2-4, generated from the wall-to-wall overlap between the test SIAM-WELD 2006 map featuring a set DictionaryOfColorNames = A = 19 spectral macro-categories and the reference NLCD 2006 map with a set LegendOfObjectClassNames = B = 16 LC classes, is guided by the inter-dictionary binary relationship R: A $\Rightarrow$ B $\subseteq$ A × B, whose entry-pair cells, shown in gray, were selected by the independent human expert (refer to Acknowledgments). Table 2-4 reveals one single systematic case of "conceptual mismatch" between the NLCD 2006 reference vegetation classes "*Scrub/Shrub*" (SS) or "*Grassland/Herbaceous*" (GH, refer to Table 2-1) and the SIAM-WELD 2006 bare soil spectral categories sbS_1, SmS_1 and aS (refer to Table 2-2). These inter-map semantic mismatches occur in geographical locations where the CONUS landscapes look like those shown in Figure 2-7. When these land surface types are observed from space with a spatial resolution of 30 m, a one-pixel surface area of 900 m$^2$ becomes a spectral mixture of sparse vegetation, rangeland, cheatgrass, dry long grass and/or short grass as foreground, with a background of sand, clay and/or rocks, especially if the percentage of vegetation cover can be slightly above the 15% of total cover required by the NLCD definitions of classes SS and GH (refer to Table 2-2). In these mixed pixels at 30 m resolution, the spectral detection of the vegetated component is impossible for a hard (crisp) classifier, while it would be more manageable by a fuzzy classifier (Baraldi 2011). In these experiments, since the SIAM expert system is run in crisp mode (refer to Section 3), then no pixel unmixing strategy can be applied to diminish or avoid the observed case of "semantic mismatch". The conclusion is that the "conceptual mismatch" between the NLCD 2006 reference vegetation classes SS and GH and the SIAM-WELD 2006 bare soil spectral categories is a possible example of systematic disagreement between the test and reference thematic maps featuring the same spatial resolution whose occurrence should be carefully scrutinized by RS experts in comparison with an "ultimate" ground truth, see Figure 2-7.

A different strategy to aesthetically (rather than formally) remove the aforementioned inter-dictionary "conceptual mismatch" would be to change color names in the SIAM color map legend, without changing the SIAM decision tree for color space hyperpolyhdralization. In other words, based on thematic evidence collected on an *a posteriori* basis from the NLCD reference map, it would be possible to change color names attached to the SIAM-WELD 2006 map legend and consider that, at the Landsat spectral and spatial resolution of an annual WELD composite of the CONUS, the SIAM spectral categories sbS_1, SmS_1 and aS are more likely to map the NLCD reference vegetation classes "*Shrub/Scrub*" (SS) or "*Grassland/herbaceous*" (GH) than bare soil surface types.

Starting from the same OAMTRX = FrequencyCount(A × B) shown in Table 2-4, two independent selections by two different RS experts of a binary relationship R: A $\Rightarrow$ B $\subseteq$ A × B provided two alternative pairs of independent O-Q$^2$I values to be jointly maximized, namely, a CVPSI2(R: A $\Rightarrow$ B) = 0.6769 with OA(OAMTRX) = 96.88% ± 0% (test case A) and a CVPSI2(R: A $\Rightarrow$ B) = 0.6804 with OA = 97.28% ± 0% with (test case B). These alternative O-$Q^2$I pairs highlight the inherent ill-posedness of any inter-dictionary conceptual harmonization, although a specific protocol to reduce heuristic decisions by human experts in the identification of a binary relationship R: A $\Rightarrow$ B $\subseteq$ A × B was proposed in the Part 1, Section 4. According to a Pareto multi-objective optimization principle, the latter O-$Q^2$I value pair should be preferred to the former. This choice proves that the OA of the test SIAM-WELD 2006 map compared with the reference NLDC 2006 map scores "very high", with a semantic information gap from sub-symbolic sensory data to symbolic NLCD classes left to be filled by further stages in the EO-IUS pipeline equal to (1 – CVPSI2) = 0.3196.

At the fine discretization level of the SIAM-WELD 2006 test map, featuring a legend A = 96 color names, another inter-map wall-to-wall overlap with the NLCD 2006 reference map, whose legend B = 16 LC classes, provided a pair of O-$Q^2$I values equal to CVPAI2(R: A $\Rightarrow$ B) = 0.5809 and OA(OAMTRX) = 95.41% ± 0% (test case C). When compared to the two pairs of O-$Q^2$I values collected from the test case A and the test case B, this third T$Q^2$I value pair proves that a finer reflectance space hyperpolyhedralization for color naming is not necessarily more convenient to cope with by human experts in the stratification of an LC classification problem according to a spectral and spatial convergence-of-evidence approach, refer to Equation (1-3) in the Part 1 (Hunt and Tyrrell 2012).

When an approximated binary relationship R: A $\Rightarrow$ B $\subseteq$ A × B was identified from set A = NLCD 16-class legend to set B = 2-level LCCS-DP 4-class legend, a binary relationship R: C $\Rightarrow$ B $\subseteq$ C × B was defined from set C = SIAM 19-class legend as rows to set B = 2-level LCCS-DP 4-class legend as columns and an OAMTRX = FrequencyCount(C × B) was generated by the wall-to-wall overlap between the test SIAM map with legend C and the reference NLCD map with legend B as reported in Table 2-5 (test case D), then O-$Q^2$I values were CVPAI2(R: C $\Rightarrow$ B) = 0.7486 and OA(OAMTRX) = 93.09% ± 0%. From these results we can infer that

> OA(Test SIAM 2006, 19 spectral macro-categories $\Rightarrow$ "Ultimate" GroundTruth 2006, 2-level LCCS-DP 4 classes) $\in$ [XX% - 6.91%, XX% + 6.91%,], where XX% = OA(Reference NLCD 2006, 2-level LCCS-DP 4 classes $\Rightarrow$ "Ultimate" GroundTruth 2006, 2-level LCCS-DP 4 classes) $\geq$ 84%, with a semantic information gap from sensory data to the 2-level LCCS-DP 4-class legend left to be filled by further stages in the EO-IUS pipeline equal to (1 – CVPAI2) = 0.2514.

This inference supports the thesis investigated by the present experimental work, where the off-the-shelf SIAM lightweight computer program for prior knowledge-based MS reflectance space hyperpolyhedralization was considered eligible for systematic Level 2 SCM product generation with an SCM legend consistent with the "augmented" 9-class LCCS-DP taxonomy.

To complete the interpretation of the OAMTRX shown in Table 2-4 two histograms of class-conditional probabilities, shown in Figure 2-8 and Fig 2-9 respectively, together with their summarized text versions, shown as Table 2-6 and Table 2-7 respectively, were generated from the OAMTRX of interest. Figure 2-8 and Table 2-6 reveal that any test SIAM-WELD 2006 spectral category conditioned by one NLCD 2006 reference class appears consistent with the NLCD class definition (refer to Table 2-1) and with *a priori* domain knowledge of RS experts about the real-world CONUS spatially sampled at 30 m resolution. Analogously, Figure 2-9 and Table 2-7 show that any NLCD 2006 reference class conditioned by one SIAM-WELD 2006 spectral category appears consistent with the spectral properties of the SIAM color type and with *a priori* domain knowledge of RS experts about the physical CONUS depicted at 30 m resolution. To conclude, class-conditional probabilities generated from Table 2-4 appear reasonable and confirm the statistical plausibility of the OAMTRX instance shown in Table 2-4 as a whole.

Figure 2-11 shows that if, for example, the boxplot of the NLCD reference class "*Developed, Open Space*" (DOS) is compared to that of reference class "*Developed, Low Intensity*" (DLI), "*Developed, Medium Intensity*" (DMI) and "*Developed, High Intensity*" (DHI, refer to Table 2-1), then a monotonic decrease of the class-conditional probability of the SIAM-WELD vegetated spectral categories conditioned by the NLCD reference class and collected at local spatial extents for a population of 86 ecoregions is observed in parallel with a monotonic increase of the class-conditional probability of the SIAM-WELD bare soil spectral categories. This is perfectly consistent with the *a priori* domain knowledge of RS experts about the spatial and spectral properties of urban and industrial area in the CONUS, in agreement with the popular vegetation-impervious surface-soil model for urban ecosystem analysis (Ridd 1995). In addition, these boxplots confirm that, at the local spatial extent of individual ecoregions, the NLCD reference classes "*Deciduous Forest*" (DF), "*Evergreen Forest*" (EF) and "*Mixed Forest*" (MF, refer to Table 2-1) are almost entirely (> 90%) covered by the SIAM-WELD vegetation spectral categories, in agreement with theoretical expectations about the SIAM-WELD test map. In line with preliminary outcomes discussed in Section 4.3 and in Figure 2-10, boxplots shown in Figure 2-11 confirm that the NLCD reference classes "*Scrub/Shrub*" (SS) and "*Grassland/Herbaceous*" (GH) have a strong heterogeneity of matches with the SIAM-WELD 2006 spectral categories collected at the ecoregion spatial extent. This is tantamount to saying that spectral signatures of these NLCD classes feature a strong variability when collected at "local" scale, also refer to Figure 2-7. More properties of the NLCD class-specific box diagrams collected at the local spatial extent of ecoregions appear reasonable, based on *a priori* human knowledge of the physical CONUS at the ecoregion spatial extent. For example, first, the NLCD reference classes "*Pasture/Hay*" (PH) and "*Cultivated Crops*" (CC, refer to Table 2-1) are largely matched across ecoregions by the SIAM-WELD vegetated spectral categories. Second, the NLCD reference class "*Perennial Ice/Snow*" (PIS, refer to Table 2-1) is best-matched across ecoregions by the SIAM-WELD spectral category MS white-as-"*Snow*" (SN, refer to Table 2-2). Third, across ecoregions, the NLCD reference class "*Open water*" (OW, refer to Table 2-1) is best-matched by the SIAM-WELD spectral category MS blue-as-"*Water or Shadow*" (WA, refer to Table 2-2). To summarize, collected at the local extent of ecoregions to account for non-stationary spatial properties, boxplots shown in Figure 2-11 are considered statistically and semantically consistent with the definitions of the two legends adopted by the test and reference maps, they agree with *a priori* domain knowledge of RS experts about the LC and LCC dynamics in the physical CONUS and appear consistent with global (non-stratified by ecoregions) statistics collected at the CONUS spatial extent reported in Section 4.2 to Section 4.4.

## 6. Conclusions

To pursue the GEO's visionary objective of a GEOSS not-yet accomplished by the RS community we advanced the following thesis: a necessary not sufficient pre-condition for a GEOSS development is the systematic generation at the ground segment of an ESA EO Level 2 product, whose general-purpose SCM legend agrees with the 3-level 9-class "augmented" LCCS-DP taxonomy (see Figure 2-2). To comply with the GEO QA4EO *Cal/Val* requirements an EO Level 2 product must be submitted to a GEO Stage 4 *Val* process by independent means. No ESA EO Level 2 product has ever been systematically generated at the ground segment. This interdisciplinary work aimed at filling an analytic and pragmatic information gap from multi-source EO big sensory data to the ESA EO Level 2 product. To fill this gap we focused our attention on a long history of prior knowledge-based MS reflectance space partitioners for static color naming, developed but never validated by space agencies, public organizations and private companies to be plugged into hybrid EO-IUSs for EO image enhancement and classification tasks in operating mode. As a potential candidate for systematic EO Level 2 SCM product generation at the ground segment to be submitted to a GEO Stage 4 *Val* we selected an off-the-shelf SIAM lightweight computer program for prior knowledge-based MS color naming, presented in the RS literature in recent years where enough information was provided for the implementation to be reproduced.

For the sake of readability this paper is split into two, the preliminary Part 1 – Theory and the present Part 2 – Validation. Original contributions of the Part 1 include, first, an eight-step protocol to identify a categorical variable-pair relationship R: A $\Rightarrow$ B from categorical variable A to categorical variable B as a hybrid combination of deductive prior beliefs with inductive evidence from data. Second, an original CVPAI2 formulation was proposed as a categorical variable-pair degree of association in a binary relationship R: A $\Rightarrow$ B. The original contribution of the present Part 2 is a novel protocol for wall-to-wall inter-map comparison without sampling, where the test and reference maps feature the same spatial resolution and spatial extent, but whose legends are not the same and must be harmonized.

Conclusions of the present Part 2 are twofold. The off-the-shelf SIAM lightweight computer program can be considered suitable for systematic generation of a Level 2 SCM product in operating mode, where the SCM legend agrees with the "augmented" 9-class LCCS-DP taxonomy. It was inferred that OA(Test SIAM 2006, 19 spectral macro-categories $\Rightarrow$ "Ultimate" GroundTruth 2006, 2-level LCCS-DP 4 classes) $\in$ [XX% - 6.91%, XX% + 6.91%,], where XX% = OA(Reference NLCD 2006, 2-level LCCS-DP 4 classes $\Rightarrow$ "Ultimate" GroundTruth 2006, 2-level LCCS-DP 4 classes) $\geq$ 84%, with a semantic information gap from sensory data to the 2-level LCCS-DP 4-class legend left to be filled by further stages in the EO-IUS pipeline equal to (1 – CVPAI2) = 0.2514 $\in$ [0, 1]. In agreement with the definition of an information processing system in operating mode proposed in Section 4, the off-the-shelf SIAM application software submitted to a Stage 4 *Val* can be considered in operating mode because its whole set of OP-$Q^2$I values scored "high".

Future developments will regard the systematic generation of a Level 2 SCM product whose map legend fully addresses the "augmented" 3-level 9-class LCCS-DP taxonomy, including quality layers such cloud and cloud-shadow. Our aim is to develop a hybrid (combined deductive and inductive) EO-IUS based on a convergence-of-evidence approach, where dominant spatial information and secondary color information are combined in line with Equation (1-3) in the Part 1. A prototypical implementation of this hybrid EO-IUS (Baraldi et al. 2016; Tiede et al. 2016) incorporates the SIAM application for color naming and exploits planar shape indexes, such as straightness-of-boundaries (Nagao and Matsuyama 1980; Soares et al. 2014), to discriminate managed (man-made) LC classes from natural surface types, such as the LCCS-DP level 3 class A11 (Cultivated and Managed Terrestrial Vegetated Areas) from class A12 (Natural and Semi-Natural Terrestrial Vegetation) and class B35

(Artificial Surfaces and Associated Areas) from class B36 (Bare Areas), see Figure 2-1.

**Acknowledgments**

To accomplish this work Andrea Baraldi was supported in part by the National Aeronautics and Space Administration (NASA) under Grant No. NNX07AV19G issued through the Earth Science Division of the Science Mission Directorate. Dirk Tiede was supported in part by the Austrian Research Promotion Agency (FFG), in the frame of project AutoSentinel2/3, ID 848009. Prof. Ralph Maughan, Idaho State University, is kindly acknowledged for his contribution as active conservationist and for his willingness to share his invaluable photo archive with the scientific community as well as the general public. Andrea Baraldi thanks Prof. Raphael Capurro, Hochschule der Medien, Germany, and Prof. Christopher Justice, Chair of the Department of Geographical Sciences, University of Maryland, for their support. Above all, the authors acknowledge the fundamental contribution of Prof. Luigi Boschetti, currently at the Department of Forest, Rangeland and Fire Sciences, University of Idaho, Moscow, Idaho, who conducted by independent means all experiments whose results are proposed in this validation paper. The authors also wish to thank the Editor-in-Chief, Associate Editor and reviewers for their competence, patience and willingness to help.

**Disclosure statement**

In accordance with XXX policy and his ethical obligation as a researcher, Andrea Baraldi reports he is the sole developer and IPR owner of the Satellite Image Automatic Mapper™ (non-registered trademark) computer program licensed to academia, public institutions and private companies, eventually free-of-charge, by the one-man-company Baraldi Consultancy in Remote Sensing that may be affected by the research reported in the enclosed paper. Andrea Baraldi has disclosed those interests fully to XXX, and he has in place an approved plan for managing any potential conflicts arising from that involvement.

**Figures and figure captions**

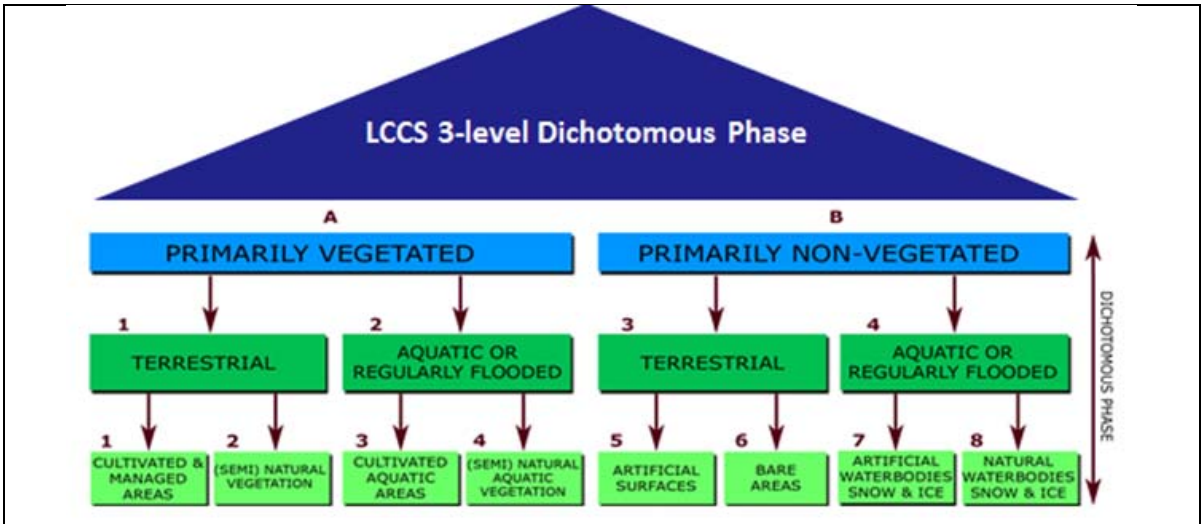

Figure 2-1. The nested 3-level LCCS-DP layers are: (i) vegetation versus non-vegetation, (ii) terrestrial versus aquatic, and (iii) managed versus natural or semi-natural. They deliver as output the following 8-class LCCS-DP taxonomy. (A11) Cultivated and Managed Terrestrial (non-aquatic) Vegetated Areas. (A12) Natural and Semi-Natural Terrestrial Vegetation. (A23) Cultivated Aquatic or Regularly Flooded Vegetated Areas. (A24) Natural and Semi-Natural Aquatic or Regularly Flooded Vegetation. (B35) Artificial Surfaces and Associated Areas. (B36) Bare Areas. (B47) Artificial Waterbodies, Snow and Ice. (B48) Natural Waterbodies, Snow and Ice. The general-purpose user- and application-independent 8-class LCCS-DP taxonomy is preliminary to a user- and application-specific LCCS Modular Hierarchical Phase (MHP) taxonomy (Di Gregorio and Jansen 2000).

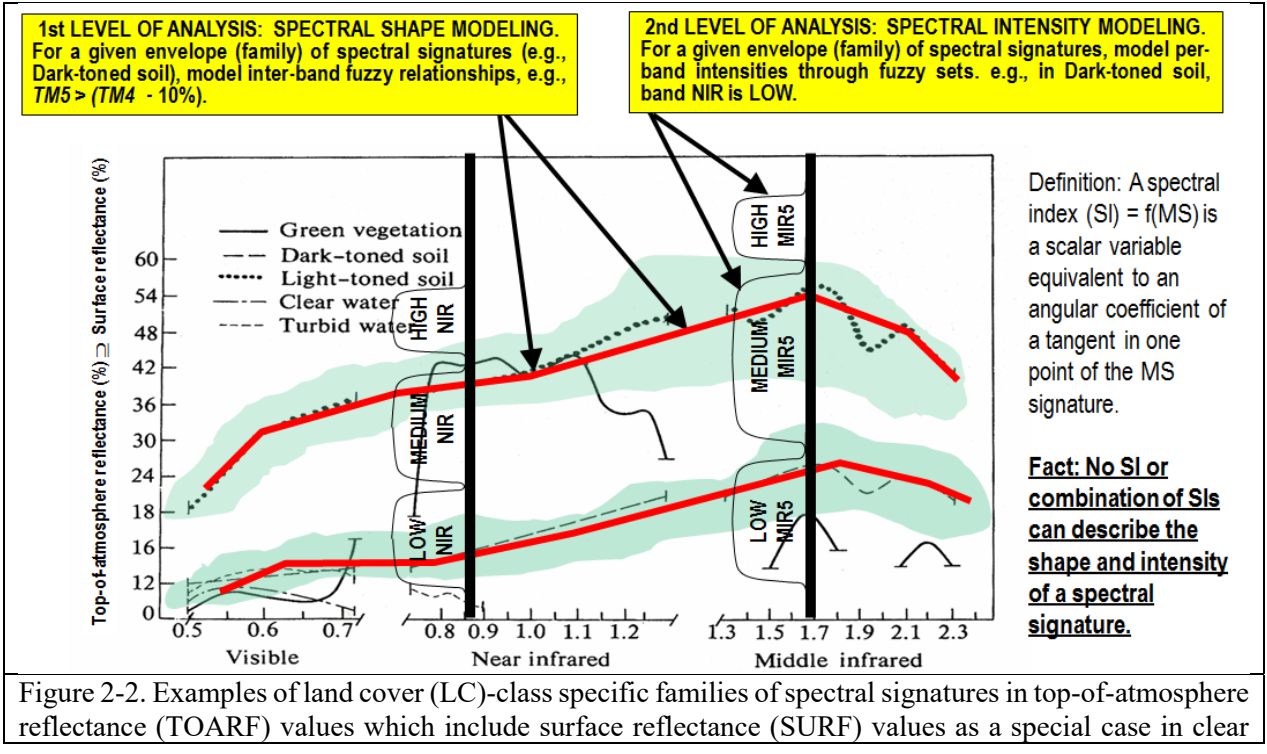

Figure 2-2. Examples of land cover (LC)-class specific families of spectral signatures in top-of-atmosphere reflectance (TOARF) values which include surface reflectance (SURF) values as a special case in clear

sky and flat terrain conditions. A within-class family of spectral signatures (e.g., dark-toned soil) in TOARF values forms a buffer zone (support area, envelope). The SIAM decision tree models each target family of spectral signatures in terms of multivariate shape and multivariate intensity as a viable alternative to multivariate analysis of spectral indexes. A typical spectral index is a scalar band ratio equivalent to an angular coefficient of a tangent in one point of the spectral signature. Infinite functions can feature the same tangent value in one point. In practice, no spectral index or combination of spectral indexes can reconstruct the multivariate shape and multivariate intensity of a spectral signature.

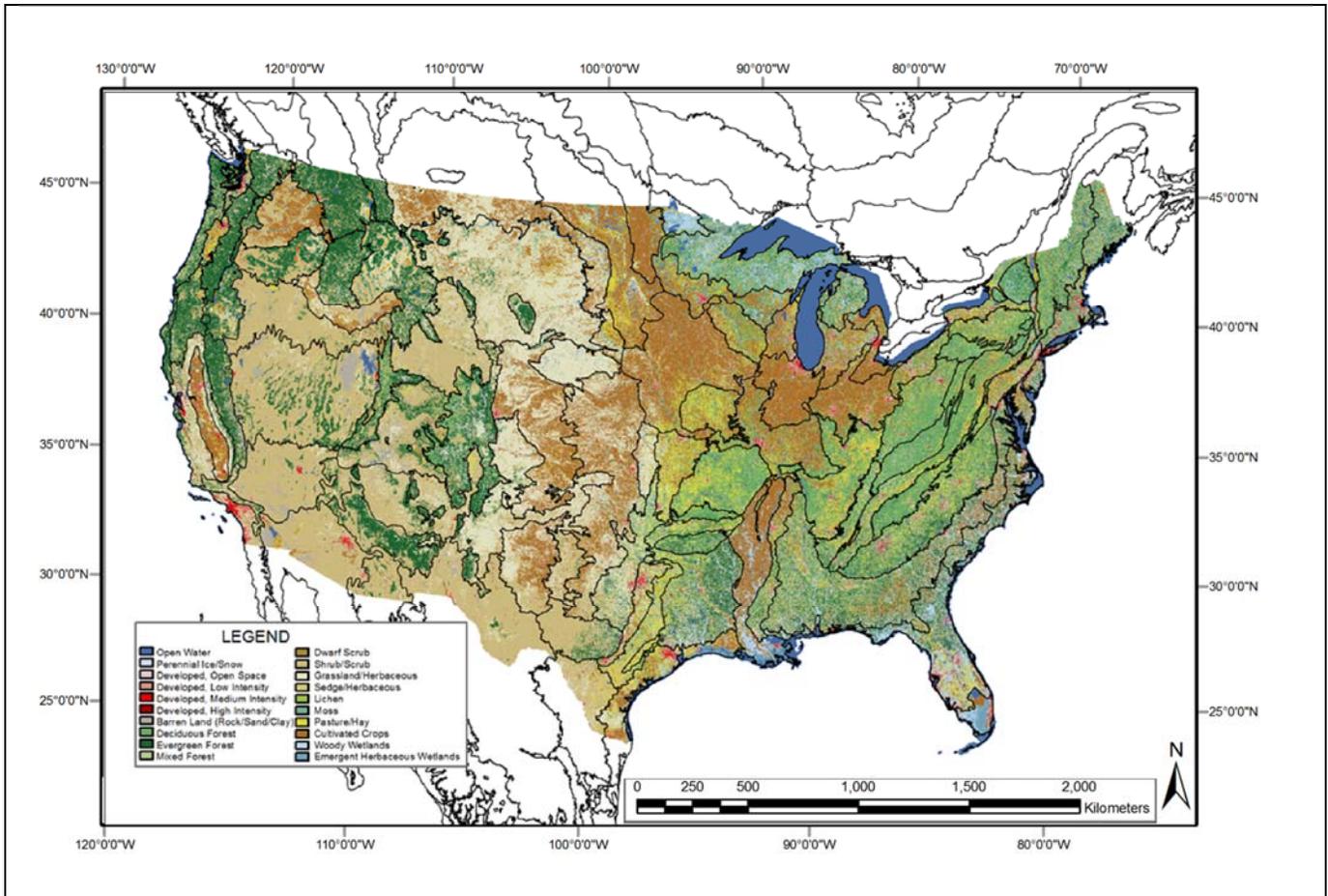

Figure 2-3. NLCD 2006 map of the CONUS. It is shown in the same scale and projection of the WELD 2006 mosaic depicted in Figure 2-4. Black lines across the NLCD 2006 map represent the boundaries of the 86 EPA Level III ecoregions of the CONUS. The NLCD 2006 map legend is shown on the left bottom side, also refer to Table 2-1.

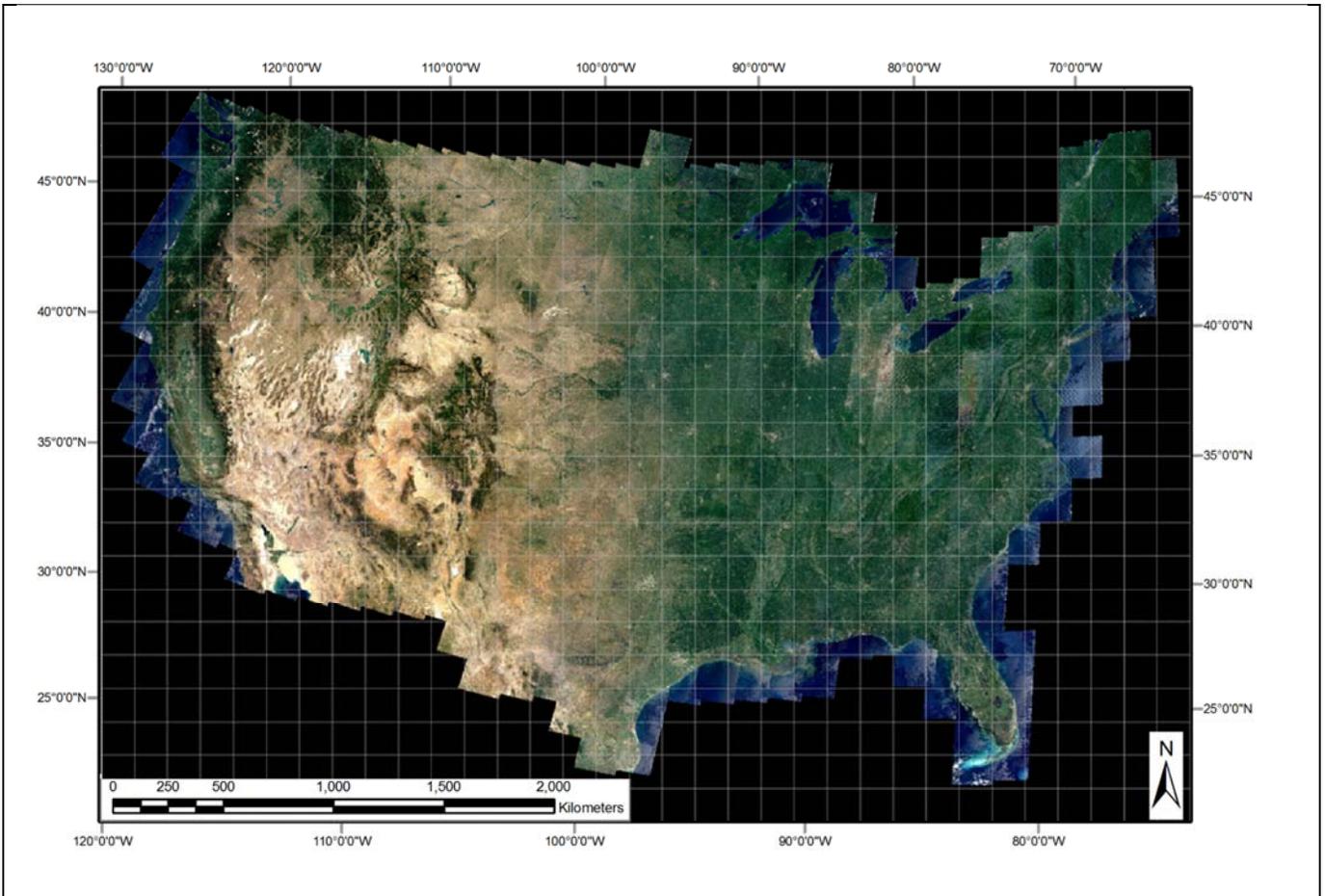

Figure 2-4. 30 m resolution annual WELD composite for the year 2006 (December 2005 to November 2006) of the conterminous U.S. (CONUS), radiometrically calibrated into top-of-atmosphere reflectance (TOARF) values. Depicted in true colors (red: Band 3, 0.63-0.69 μm; green: Band 2, 0.53-0.61 μm, and blue: Band 1, 0.45-0.52 μm). The white grid shows locations of the 501 WELD tiles of the CONUS. Each tile is 5000×5000 pixels in size, covering a surface area of 150×150 km. Pixels are geographically projected in the Albers Equal Area projection.

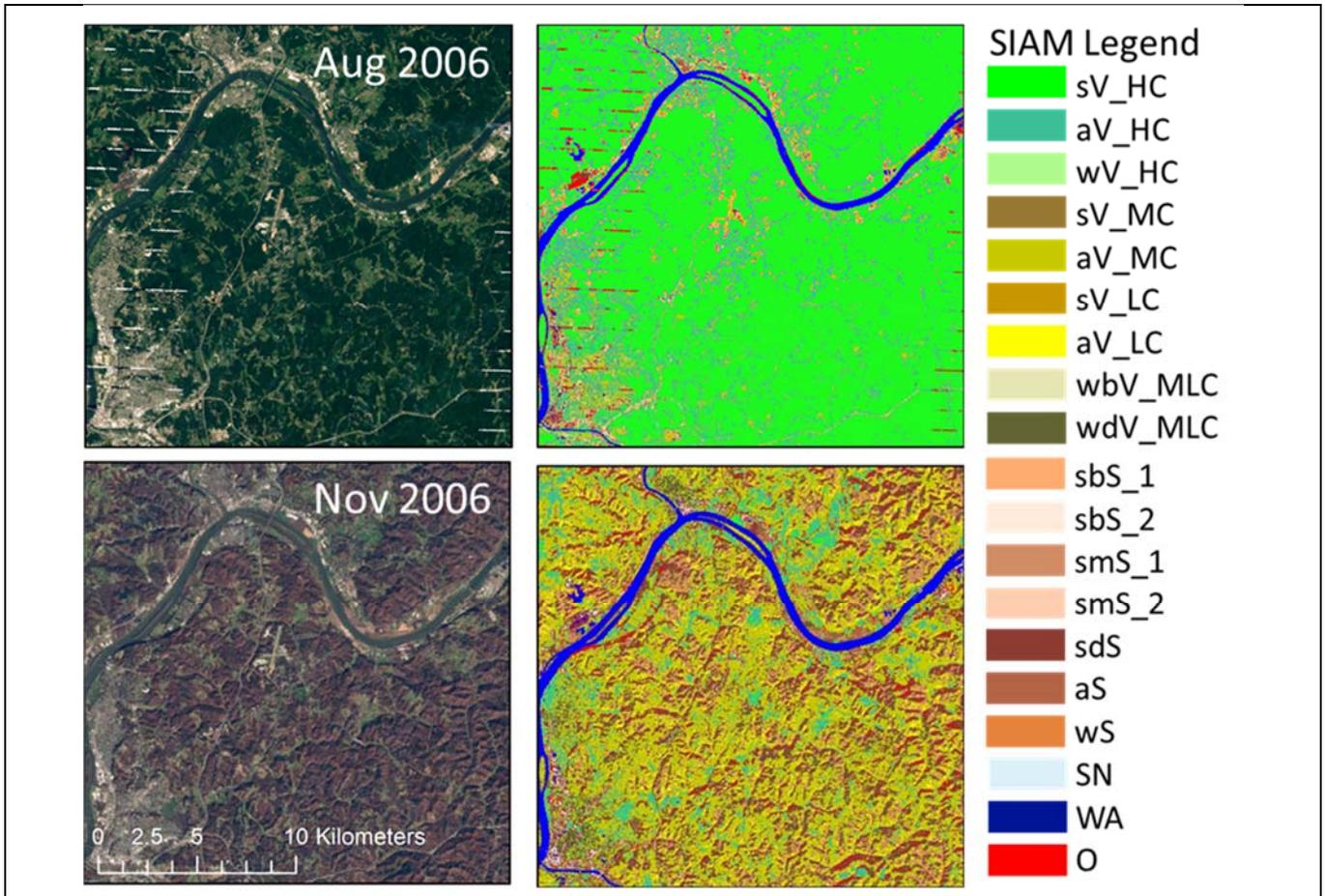

Figure 2-5. Changes in the SIAM's spectral macro-category labeling due to vegetation phenology affecting the monthly WELD composite. *Left side*: 30 m resolution monthly WELD composites, radiometrically calibrated into top-of-atmosphere reflectance (TOARF) values, for August and November 2006, showing an area predominantly covered by broadleaf forest in the Mid-Western United States (Ohio). Depicted in true colors (red: Band 3, 0.63-0.69 μm; green: Band 2, 0.53-0.61 μm, and blue: Band 1, 0.45-0.52 μm). To allow inter-image comparison, the two images are displayed with an identical contrast stretch. *Right side:* SIAM-WELD color maps generated from the two WELD images shown on the left side. The SIAM map legend, consisting of 19 spectral macro-categories, is shown on the right side, also refer to Table 2-2.

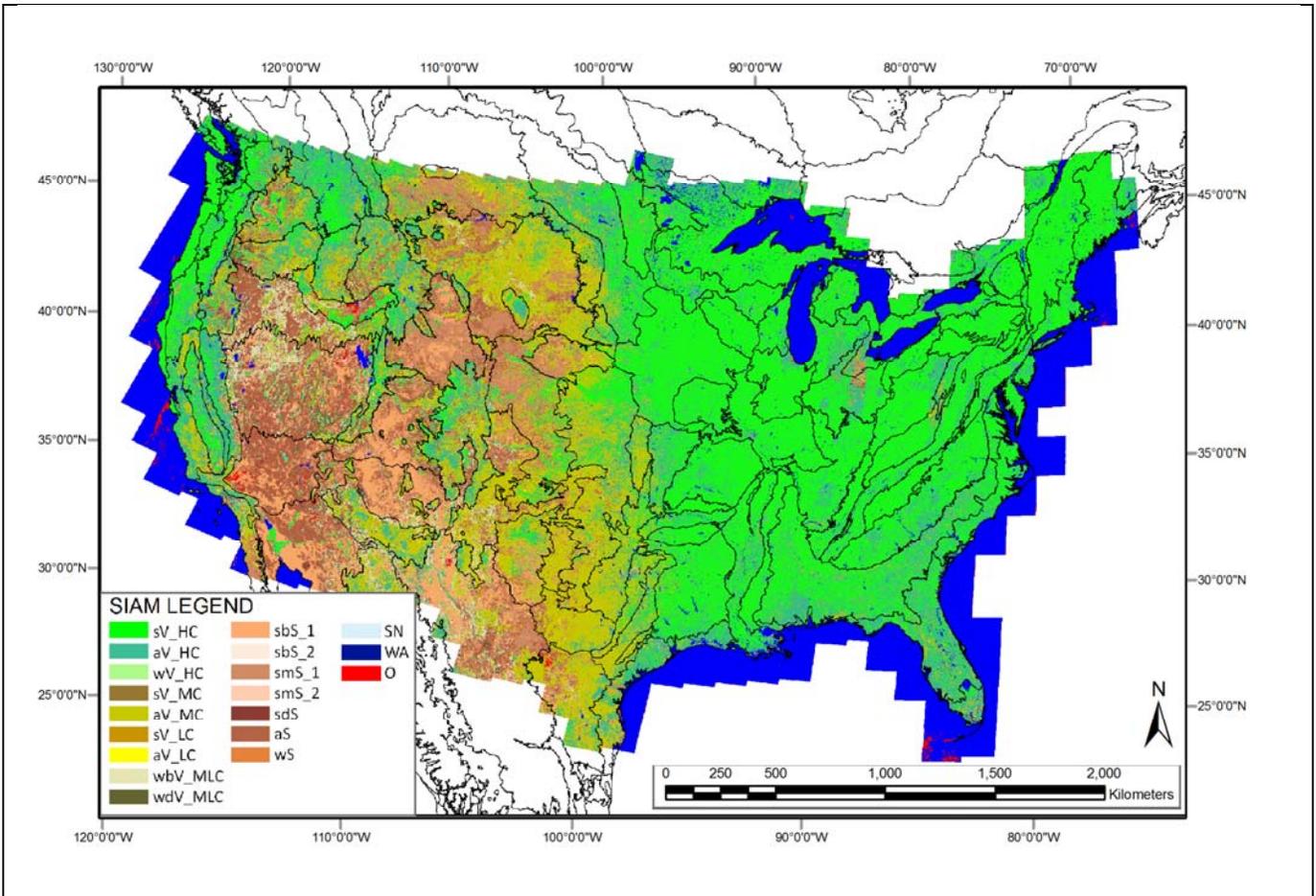

Figure 2-6. SIAM-WELD 2006 color map at an intermediate discretization level of 48 color names, reassembled into 19 spectral macro-categories. Black lines across the SIAM-WELD 2006 map represent the boundaries of the 86 EPA Level III ecoregions of the CONUS. The SIAM map legend is shown on the left bottom corner, also refer to Table 2-2.

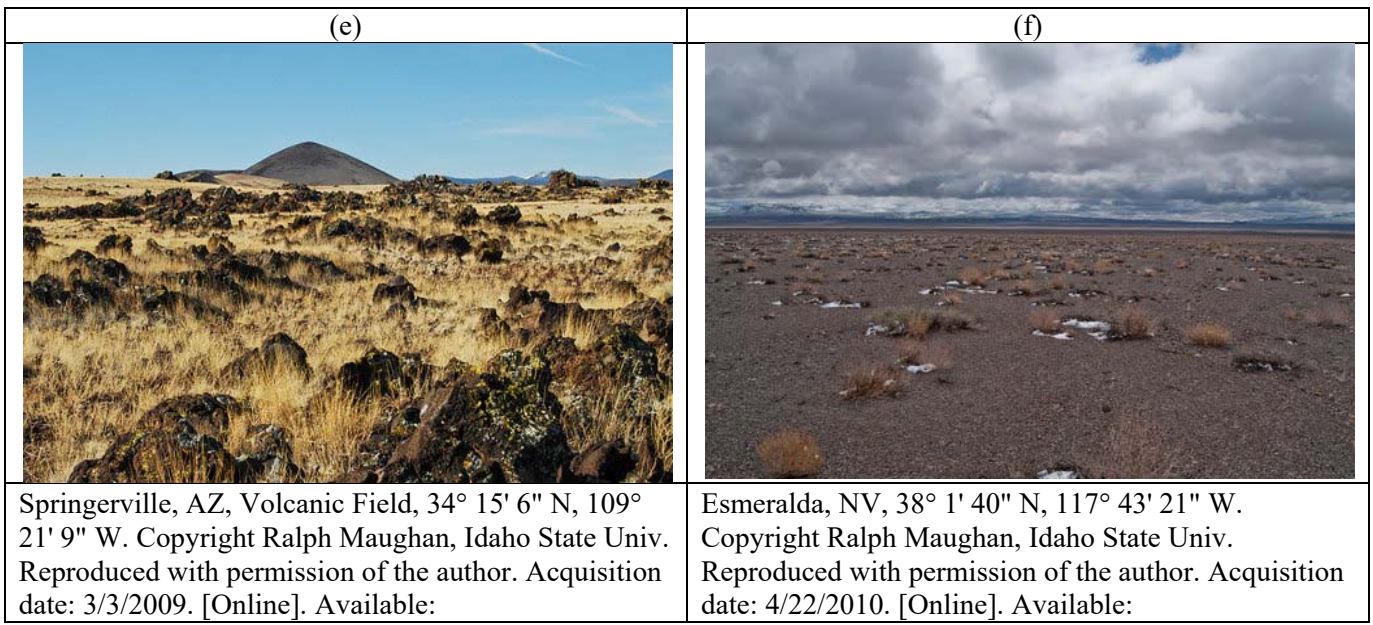

| (e) | (f) |
|---|---|
| Springerville, AZ, Volcanic Field, 34° 15' 6" N, 109° 21' 9" W. Copyright Ralph Maughan, Idaho State Univ. Reproduced with permission of the author. Acquisition date: 3/3/2009. [Online]. Available: | Esmeralda, NV, 38° 1' 40" N, 117° 43' 21" W. Copyright Ralph Maughan, Idaho State Univ. Reproduced with permission of the author. Acquisition date: 4/22/2010. [Online]. Available: |

| http://www.panoramio.com (accessed on 24 Feb. 2013). | http://www.panoramio.com (accessed on 24 Feb. 2013). |
|---|---|
| (c) | (d) |
| 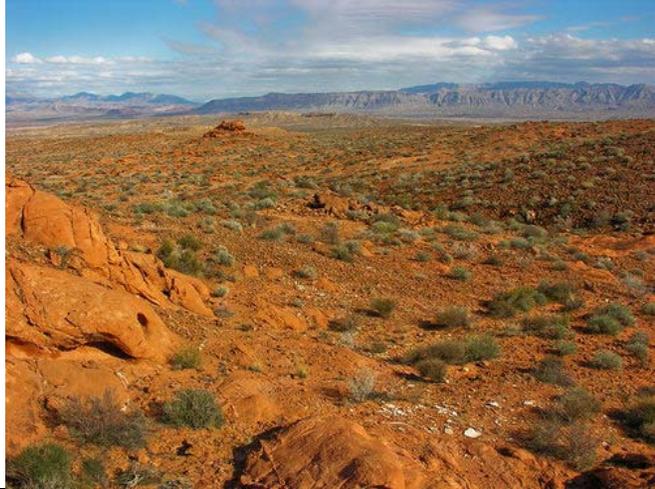 | 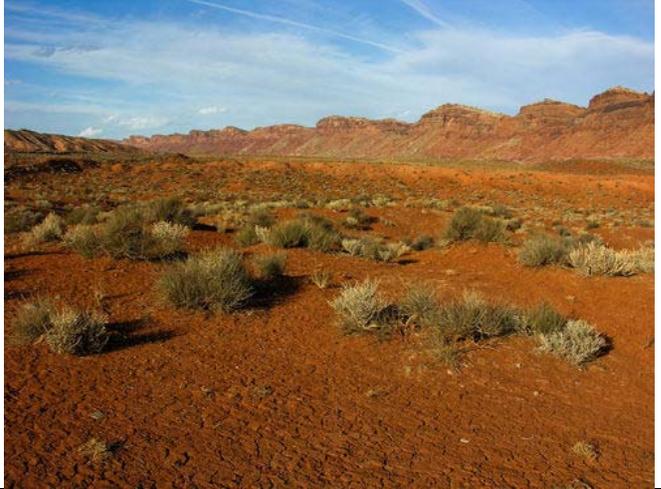 |
| Overton, NV, 36° 25' 42" N, 114° 27' 21" W. Copyright Ralph Maughan, Idaho State Univ. Reproduced with permission of the author. Acquisition date: 2/11/2009. [Online]. Available: http://www.panoramio.com (accessed on 24 Feb. 2013). | San Juan, UT, 37° 16' 43" N, 109° 40' 27" W. Copyright Ralph Maughan, Idaho State Univ. Reproduced with permission of the author. Acquisition date: 3/4/2009. [Online]. Available: http://www.panoramio.com (accessed on 24 Feb. 2013). |
| (a) | (b) |
| 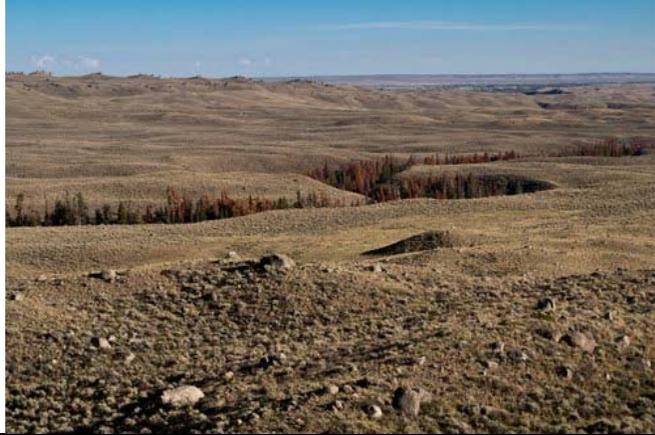 | 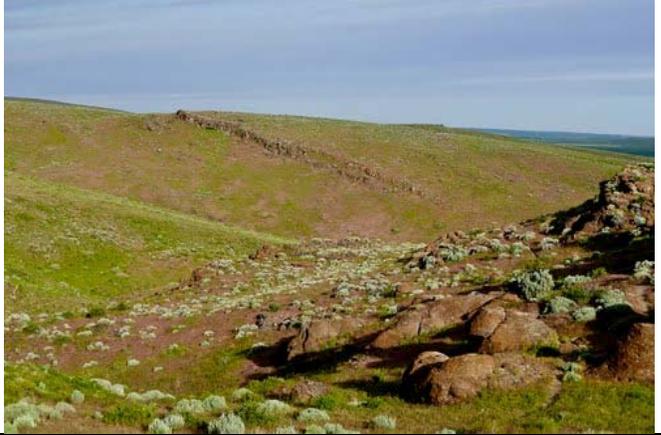 |
| Sublette, WY, Rangeland, 42° 51' 37" N, 109° 43' 7" W. Copyright Ralph Maughan, Idaho State Univ. Reproduced with permission of the author. Acquisition date: 6/16/2011. [Online]. Available: http://www.panoramio.com (accessed on 24 Feb. 2013). | Twin Falls, ID, Ripening cheatgrass infestation, 42° 23' 52" N, 114° 21' 9" W. Copyright Ralph Maughan, Idaho State Univ. Reproduced with permission of the author. Acquisition date: April 2010? [Online]. Available: http://www.panoramio.com (accessed on 24 Feb. 2013). |

Figure 2-7. Examples of geographic locations mapped as vegetation classes "*Scrub/Shrub*" (SS) or "*Grassland/Herbaceous*" (GH) in the NLCD 2006 reference map (refer to Table 2-1) and predominantly as bare soil spectral categories (sbS_1, SmS_1, aS) in the SIAM-WELD 2006 test map (refer to Table 2-2), as pointed out in Table 2-8. The SIAM's color names sbS_1, SmS_1 and aS mean that, from space, with a pixel size of 30 m × 30 m = 900 m$^2$, the contribution of sparse vegetation, rangeland, cheatgrass, dry long grass or short grass as foreground, mixed with a background of sand, clay or rocks, like those shown in these pictures, becomes extremely difficult to detect, especially if a hard (crisp, defuzzified) label rather than a set of fuzzy class labels must be provided as the output product.

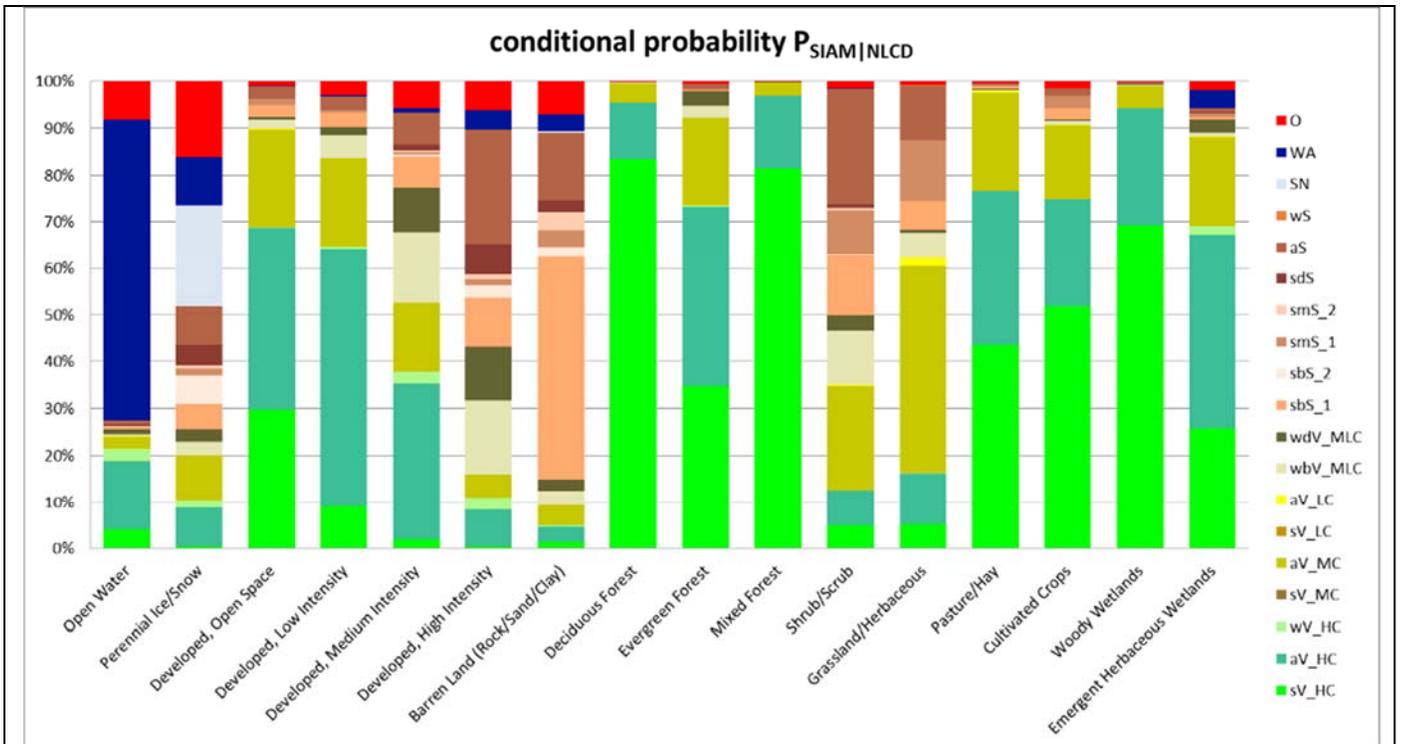

Figure 2-8. Histogram of the conditional probabilities of the 19 SIAM-WELD 2006 spectral macro-categories (shown as the right column of acronyms, refer to Table 2-2) at the intermediate color discretization level, conditioned by one-of-16 NLCD 2006 classes (listed along the horizontal axis). These conditional probabilities are derived from Table 2-4 by normalizing each cell of Table 2-4 by its column-sum. The same conditional probabilities are summarized in text form in Table 2-6. In this histogram, pseudo-colors associated with the SIAM color types make the interpretation of the histogram bins more intuitive. Green pseudo-colors are associated with vegetation spectral categories (see labels of type xV_x on the right column of labels), brown pseudo-colors are selected for bare soil spectral categories (see labels of type xS_x on the right column of labels), the pseudo-color blue is chosen for the "*Water or Shadow*" (WA) spectral category, the light blue pseudo-color is linked to the snow (SN) spectral category etc. As a consequence, the bin of the NLCD class "*Open Water*" is expected to look blue, bins of the NLCD vegetation classes are expected to look green, etc.

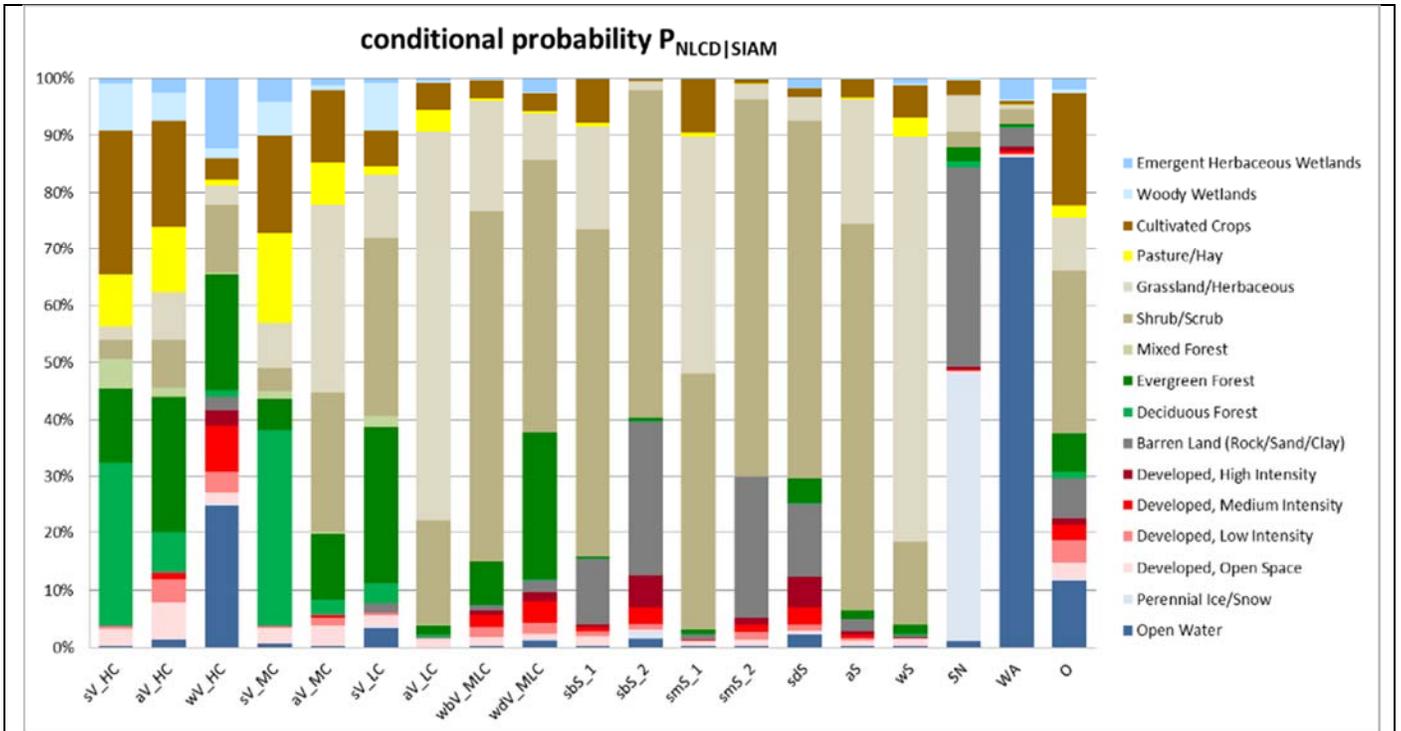

Figure 2-9. Histogram of the conditional probabilities of the 16 NLCD 2006 classes (shown as the right column of class names) conditioned by one-of-19 SIAM-WELD 2006 spectral macro-categories (listed along the horizontal axis as acronyms, refer to Table 2-2) at the intermediate color discretization level. This histogram is derived from Table 2-4 by normalizing each cell of Table 2-4 by its row-sum. The same conditional probabilities are summarized in text form in Table 2-7. In this histogram, pseudo-colors associated with the NLCD classes make the interpretation of the histogram bins more intuitive. Green pseudo-colors are associated with vegetation NLCD classes, brown pseudo-colors are selected for bare soil NLCD classes, the pseudo-color blue is chosen for the NLCD class "*Open Water*", the light blue pseudo-color is linked to the NLCD class "*Perennial Ice/Snow*", etc. As a consequence, the bin of the SIAM's spectral category "*Water or Shadow*" (WA) is expected to look blue, bins of the SIAM's vegetation spectral categories (see labels of type xV_x along the horizontal axis) are expected to look green, etc.

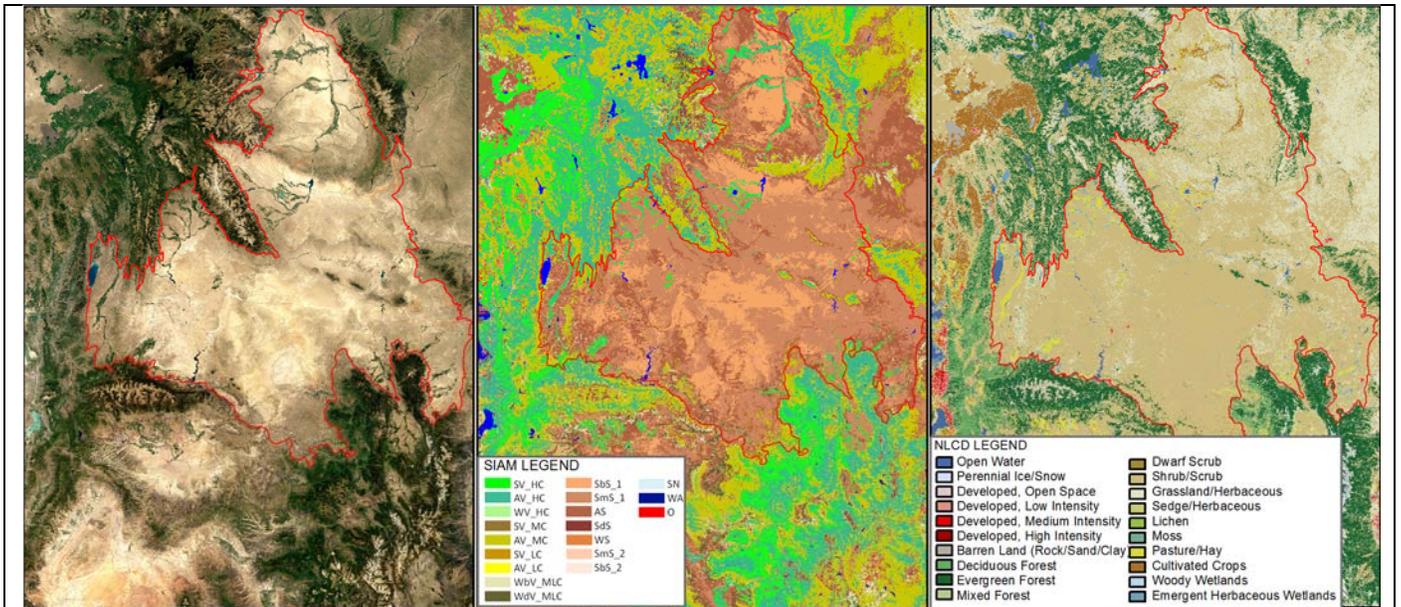

Figure 2-10. Wyoming Basin ecoregion, as part of the "North American deserts" level 1 ecoregion 10.1.4. *Left*: WELD 2006 tile (true color). *Middle*: SIAM test map of the WELD 2006 tile shown at left, with 19 spectral macro-categories at the intermediate color discretization level. *Right*: NLCD 2006 reference map, featuring 16 LC classes. In these three images, the boundary of the Wyoming Basin ecoregion is overlaid in red. The desertic Wyoming Basin ecoregion is classified as predominantly "*Scrub/Shrub*" (SS) and "*Grassland/Herbaceous*" (GH) in the NLCD 2006 reference map (refer to Table 2-1), and predominantly as bare soil (sbS_1, SmS_1, aS) in the SIAM-WELD 2006 test map (refer to Table 2-2). This phenomenon of comprehensive "semantic mismatch" between the NLCD 2006 and SIAM-WELD 2006 thematic maps is explained thoroughly in Section 5.C.

(a)

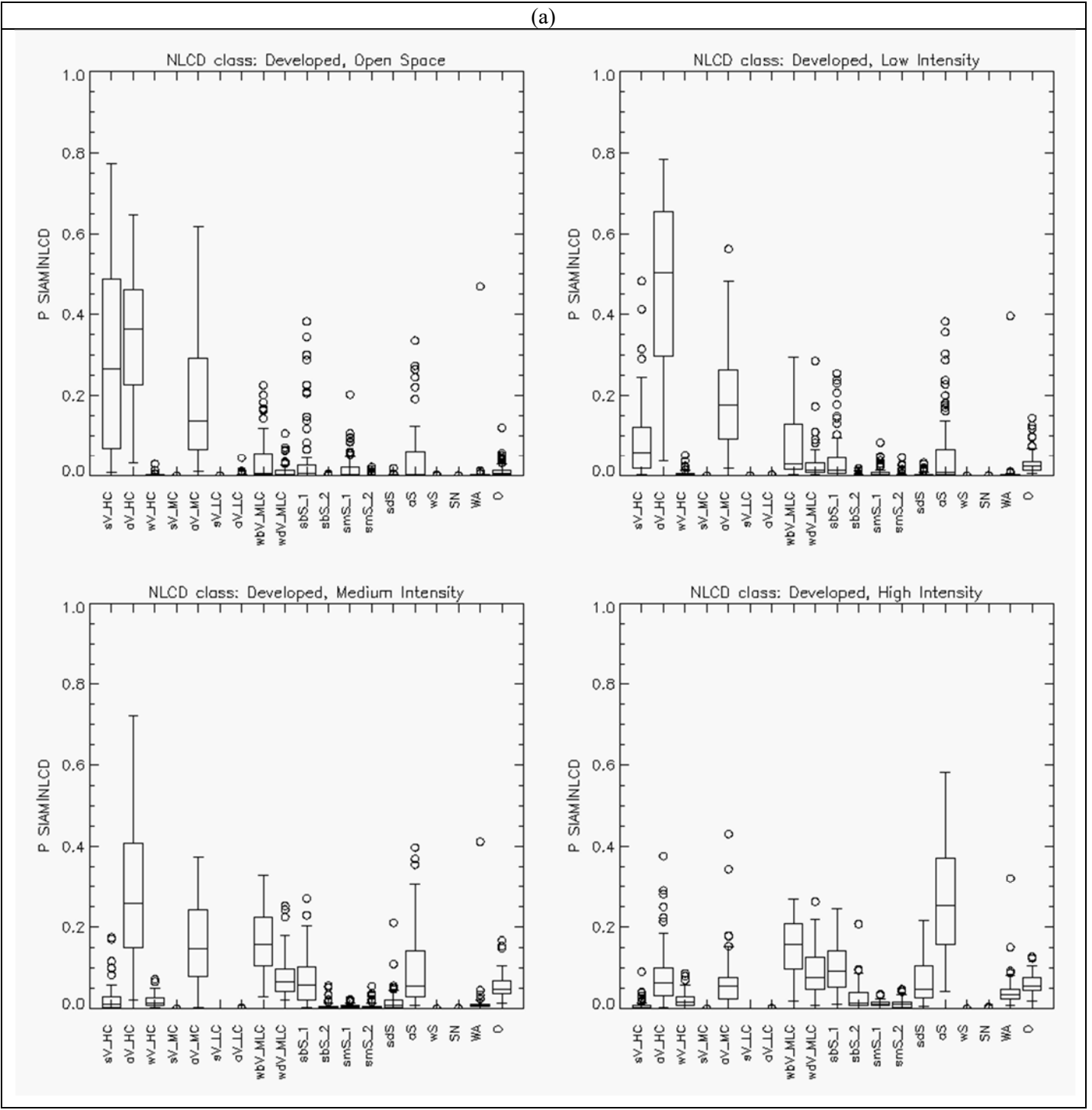

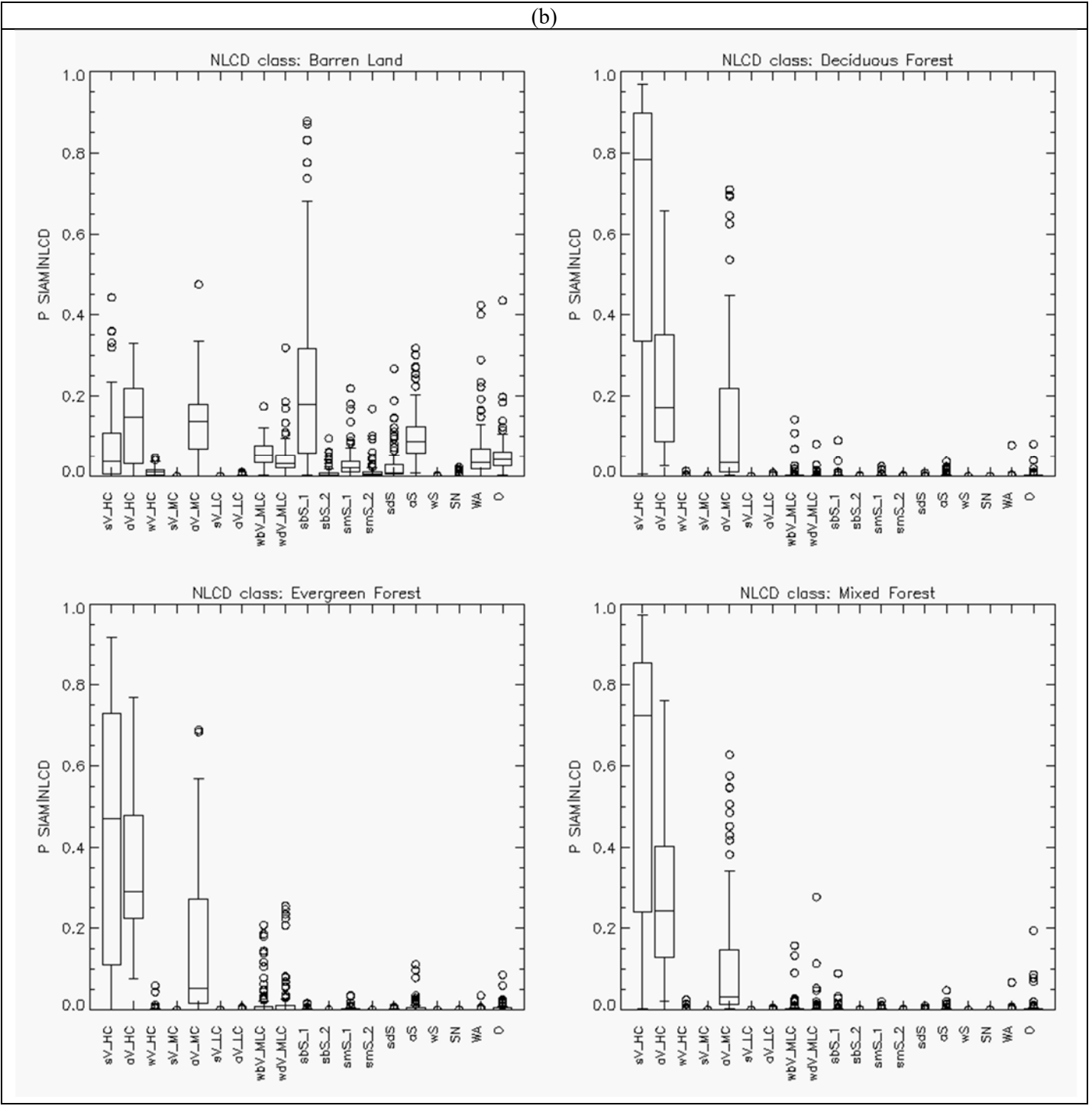

(c)

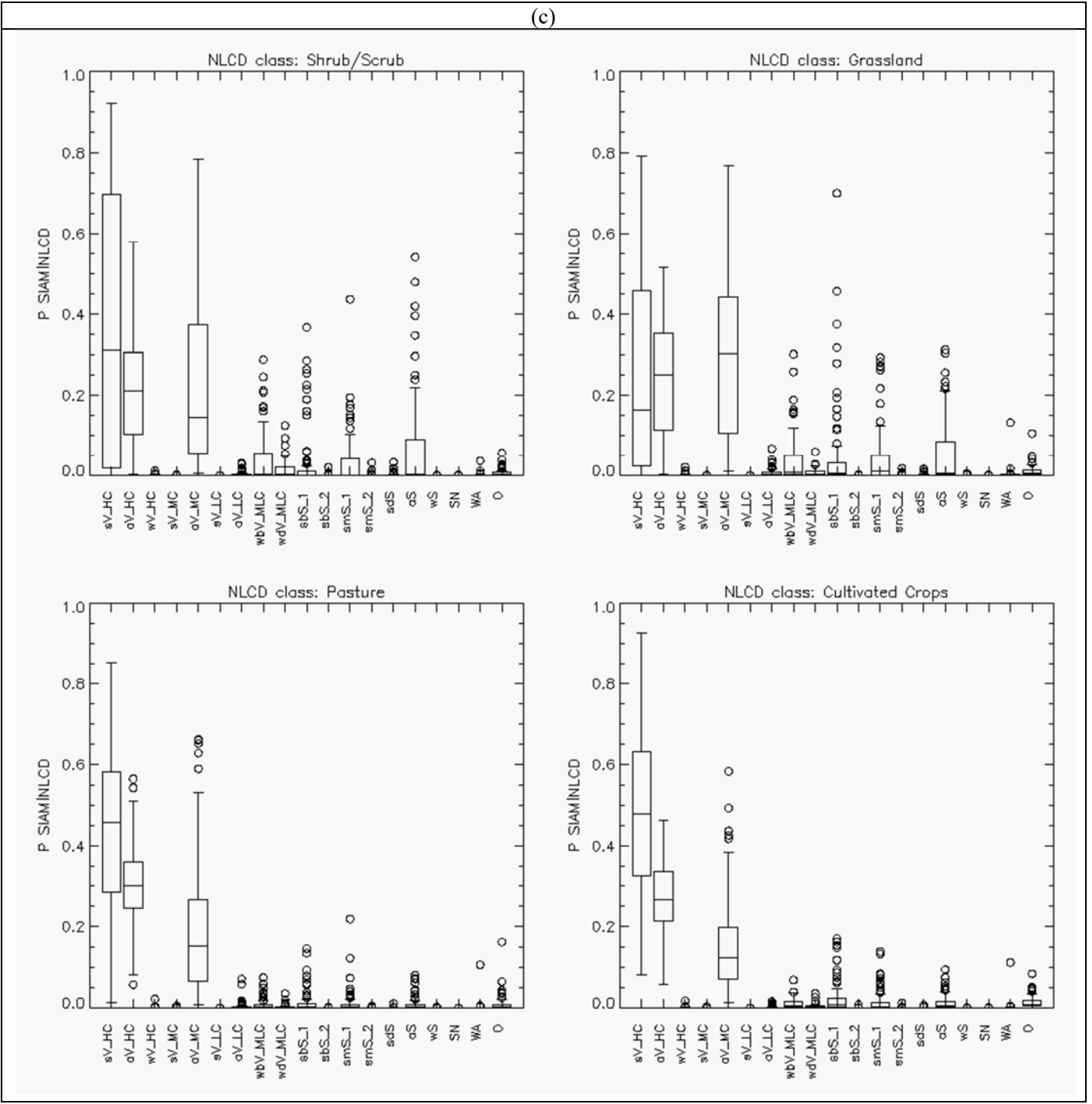

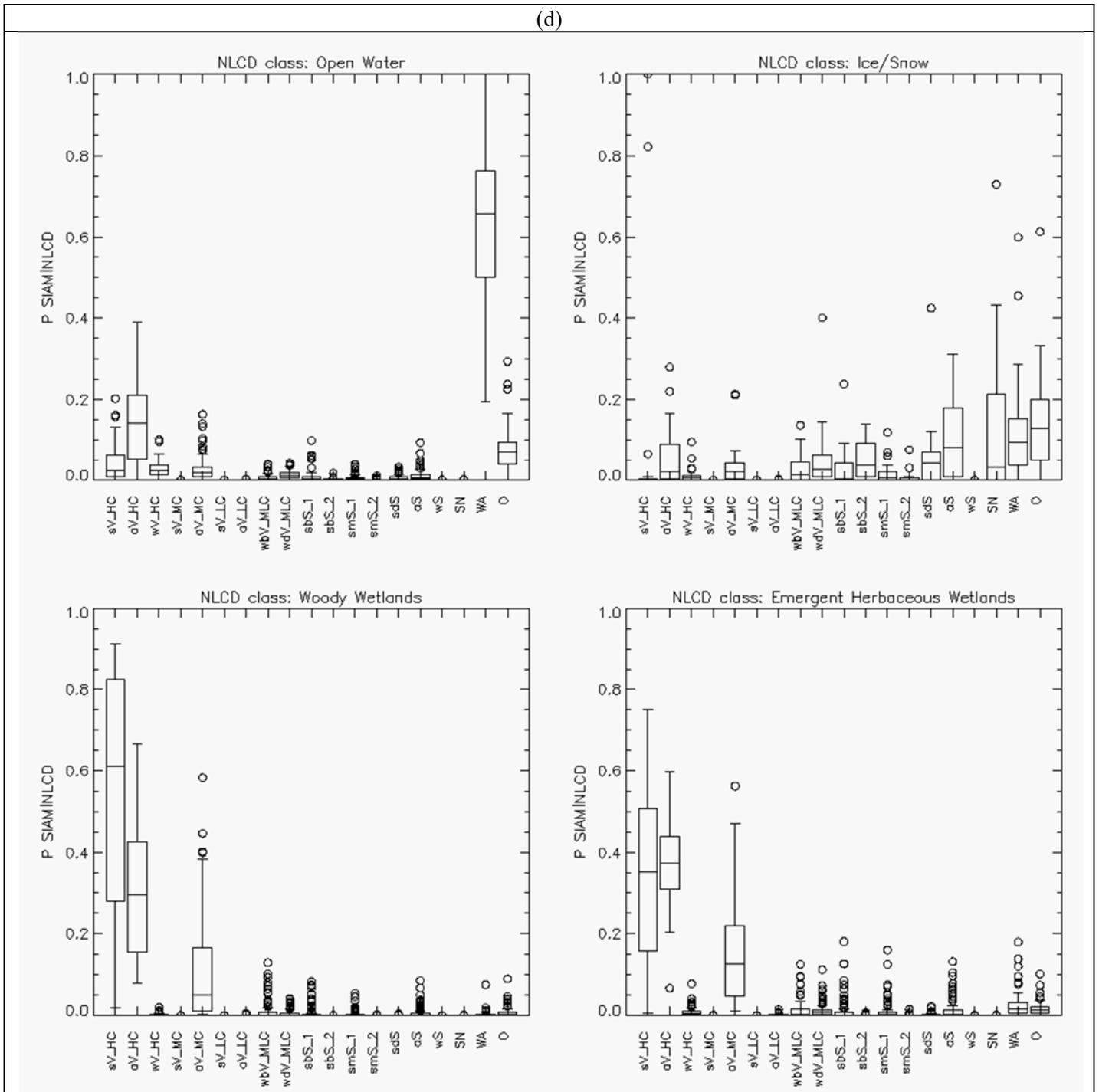

Figure 2-11(a) - (d). Reference NLCD class-specific box-and-whisker diagrams, identified by index $r = 1, ..., RC = 16$, of the NLCD class-conditional probabilities $p(\text{SIAM-WELD}_{er,\,t} \,/\, \text{NLCD}_{er,\,r})$, with $t = 1, ..., TC = 19$, collected across ecoregions $er = 1, ..., ER = 86$. The 19 spectral categories of the SIAM-WELD test map, identified by their acronyms (refer to Table 2-2), are distributed along the x axis of each NLCD class-specific diagram. Each of the 19 boxes in a box-and-whisker diagram extends from the 25th to the 75th percentile, with a horizontal line to represent the median (50th percentile) of the distribution. The whiskers extend to the maximum or minimum value of the data set, or to 1.5 times the interquantile range, whichever comes first. If there is data beyond this range, it is represented by open circles.

**Tables and table captions**

| NLCD 2001/2006/2011 Classification Scheme (Legend), Level II | | | | LCCS-DP, level 1: A = Veg, B = Non-Veg, and level 2: 1 = Terrestrial, 2 = Aquatic |
|---|---|---|---|---|
| Code | ID | Name | Land cover (LC) Class Definition | ID |
| 11 | OW | Open water | OW: Areas of open water, generally with less than 25% cover of vegetation or soil | B4 - Non-vegetated aquatic |
| 12 | PIS | Perennial Ice/Snow | PIS: Areas characterized by a perennial cover of ice and/or snow, generally greater than 25% of total cover. | B4 |
| 21<br>22<br>23<br>24 | DOS<br>DLI<br>DMI<br>DHI | • Developed, Open Space<br>• Developed, Low Intensity<br>• Developed, Medium Intensity<br>• Developed, High Intensity | DOS: Includes areas with a mixture of some constructed materials, but mostly vegetation in the form of lawn grasses. Impervious surfaces account for less than 20 percent of total cover. These areas most commonly include large-lot single-family housing units, parks, golf courses, and vegetation planted in developed settings for recreation, erosion control, or aesthetic purposes.<br>DLI, DMI, DHI: refer to the "National Land Cover Database 2006 (NLCD2006)," Multi-Resolution Land Characteristics Consortium (MRLC), 2013. | B3 - Non-vegetated terrestrial / A1 - Vegetated terrestrial |
| 31 | BL | Barren Land (Rock/ Sand/ Clay) | BL: Barren areas of bedrock, desert pavement, scarps, talus, slides, volcanic material, glacial debris, sand dunes, strip mines, gravel pits and other accumulations of earthen material. Generally, vegetation accounts for less than 15% of total cover. As a consequence of this constraint, class BL covers only 1.21% of the CONUS total surface. | B3 |
| 41<br>42<br>43 | DF<br>EF<br>MF | • Deciduous Forest<br>• Evergreen Forest<br>• Mixed Forest | DF: Areas dominated by trees generally greater than 5 meters tall, and greater than 20% of total vegetation cover. More than 75 percent of the tree species shed foliage simultaneously in response to seasonal change.<br>EF: Areas dominated by trees generally greater than 5 meters tall, and greater than 20% of total vegetation cover. More than 75 percent of the tree species maintain their leaves all year. Canopy is never without green foliage.<br>MF: Mixed Forest - Areas dominated by trees generally greater than 5 meters tall, and greater than 20% of total vegetation cover. Neither deciduous nor evergreen species are greater than 75 percent of total tree cover. | A1 |
| 51<br>52 | -<br>SS | • Dwarf Scrub [2]<br>• Scrub/Shrub | SS: Areas dominated by shrubs; less than 5 meters tall with shrub canopy typically greater than 20% of total vegetation. This class includes true shrubs, young trees in an early successional stage or trees stunted from environmental conditions. The aforementioned definition of class BL means that class SS may feature a vegetated cover which accounts for 15% of total cover or more. | A1/ B3 |

| | | | | | |
|---|---|---|---|---|---|
| 71 72 73 74 | GH - - - | • Grassland/Herbaceous<br>• Sedge Herbaceous [2]<br>• Lichens [2]<br>• Moss [2] | GH: Areas dominated by grammanoid or herbaceous vegetation, generally greater than 80% of total vegetation. These areas are not subject to intensive management such as tilling, but can be utilized for grazing. The aforementioned definition of class BL means that class GH may feature a vegetated cover which accounts for 15% of total cover or more. | | A1/B3 |
| 81 82 | PH CC | • Pasture/Hay<br>• Cultivated Crops | PH: Areas of grasses, legumes, or grass-legume mixtures planted for livestock grazing or the production of seed or hay crops, typically on a perennial cycle. Pasture/hay vegetation accounts for greater than 20 percent of total vegetation.<br>CC: Areas used for the production of annual crops, such as corn, soybeans, vegetables, tobacco, and cotton, and also perennial woody crops such as orchards and vineyards. Crop vegetation accounts for greater than 20% of total vegetation. This class also includes all land being actively tilled. | | A1 |
| 90 95 | WW EHW | • Woody Wetlands<br>• Emergent Herbaceous Wetland | WW: Areas where forest or shrubland vegetation accounts for greater than 20 percent of vegetative cover and the soil or substrate is periodically saturated with or covered with water.<br>EHW: Areas where perennial herbaceous vegetation accounts for greater than 80% of vegetative cover and the soil or substrate is periodically saturated with or covered with water. | | A2 – Vegetated aquatic |

Table 2-1. Definition of the NLCD 2001/2006/2011 classification taxonomy, Level II. [2]Alaska only. For further details, refer to the "National Land Cover Database 2006 (NLCD2006)," Multi-Resolution Land Characteristics Consortium (MRLC), 2013. The right column instantiates a possible binary relationship R: A $\Rightarrow$ B $\subseteq$ A × B from set A = NLCD legend to set B = 2-level 4-class Dichotomous Phase (DP) taxonomy of the Food and Agriculture Organization of the United Nations (FAO) - Land Cover Classification System (LCCS) (Di Gregorio and Jansen 2000), refer to Figure 2-1.

| SIAM, Intermediate discretization level (featuring 48 spectral categories) reassembled into 19 spectral macro-categories | | | | LCCS-DP, level 1: A = Veg, B = Non-Veg, and level 2: 1 = Terrestrial, 2 = Aquatic |
|---|---|---|---|---|
| ID | Abbreviation | OR-Aggregations | Spectral macro-category name | ID |
| 1 | sV_HC | 1 | Strong evidence vegetation, high canopy cover | A1 - Vegetated terrestrial |
| 2 | aV_HC | 1 | Average evidence vegetation, high canopy cover | A1 |
| 3 | wV_HC | 1 | Weak evidence vegetation, high canopy cover | A1 |
| 4 | sV_MC | 1 | Strong evidence vegetation, medium canopy cover | A1 |
| 5 | aV_MC | 1 | Average evidence vegetation, medium canopy cover | A1 |
| 6 | sV_LC | 1 | Strong evidence vegetation, low canopy cover | A1 |
| 7 | aV_LC | 1 | Average evidence vegetation, low canopy cover | A1 |

| 8 | wbV_MLC | 1 | Weak evidence bright vegetation, medium or low canopy cover | A1 |
| 9 | wdV_MLC | 1 | Weak evidence dark vegetation, medium or low canopy cover | A1/A2 - Vegetated aquatic |
| 10 | sbS_1 | 1 | Strong evidence bright soil AND NIR <= MIR | B3 - Non-vegetated terrestrial |
| 11 | sbS_2 | 1 | Strong evidence bright soil AND NIR > MIR | B3 |
| 12 | smS_1 | 1 | Strong evidence medium soil AND NIR <= MIR | B3 |
| 13 | smS_2 | 1 | Strong evidence medium soil AND NIR > MIR | B3 |
| 14 | sdS | 1 | Strong evidence dark soil | B3 |
| 15 | aS | 1 | Average evidence soil | B3 |
| 16 | wS | 1 | Weak evidence soil | B3 |
| 17 | SN | 2 | Snow | B4 - Non-vegetated aquatic |
| 18 | WA | 6 | Water or Shadow | B4 |
| 19 | O | 24 | Others | B - Non-vegetated |
|  | **TOT.** | 48 |  |  |

Table 2-2. List of the 19 spectral macro-categories generated from the aggregation of the SIAM's 48 spectral categories originally detected at the intermediate level of color quantization. The "*Water or Shadow*" (WA) spectral macro-category results from the aggregation of six original SIAM categories, the "*Snow*" (SN) spectral macro-category from two and the spectral macro-category "*Others*" (O) from the aggregation of 24 original spectral categories covering disturbances typically minimized or removed in an annual composite (clouds, smoke plumes, fire fronts, etc.) as well as the original spectral category "*Unknowns*". Hence, (19 - 3) + 6 + 2 + 24 = 48, which is the SIAM's intermediate discretization level. In the proposed names of spectral macro-categories, acronym Near Infra-Red (NIR) indicates Landsat TM/ETM+ band 4 (0.9 μm) and Medium Infra-Red (MIR) indicates Landsat TM/ETM+ band 5 (1.6 μm). The right column instantiates a possible binary relationship R: A $\Rightarrow$ B $\subseteq$ A × B from set A = SIAM legend to set B = 2-level 4-class Dichotomous Phase (DP) taxonomy of the Food and Agriculture Organization of the United Nations (FAO) - Land Cover Classification System (LCCS) (Di Gregorio and Jansen 2000), refer to Figure 2-1.

| **Spectral Category** | 2006 | 2007 | 2008 | 2009 | Mean | Std Dev |
|---|---|---|---|---|---|---|
| sV_HC | 33.11% | 32.56% | 33.79% | 34.06% | 33.38% | 0.68% |
| aV_HC | 19.94% | 23.31% | 20.02% | 20.86% | 21.03% | 1.57% |
| wV_HC | 0.18% | 0.17% | 0.17% | 0.19% | 0.18% | 0.01% |
| sV_MC | 0.00% | 0.00% | 0.00% | 0.00% | 0.00% | 0.00% |
| aV_MC | 20.05% | 18.79% | 18.07% | 17.93% | 18.71% | 0.97% |
| sV_LC | 0.00% | 0.00% | 0.00% | 0.00% | 0.00% | 0.00% |
| aV_LC | 0.40% | 0.18% | 0.31% | 0.22% | 0.28% | 0.10% |
| wbV_MLC | 4.12% | 3.60% | 3.73% | 3.30% | 3.69% | 0.34% |
| wdV_MLC | 1.48% | 1.37% | 1.19% | 1.53% | 1.39% | 0.15% |
| *Total vegetation* | *79.28%* | *79.98%* | *77.29%* | *78.10%* | *78.66%* | *1.20%* |
| sbS_1 | 5.00% | 5.44% | 6.28% | 5.39% | 5.53% | 0.54% |
| sbS_2 | 0.09% | 0.13% | 0.08% | 0.12% | 0.11% | 0.02% |
| smS_1 | 4.65% | 3.51% | 5.38% | 4.78% | 4.58% | 0.78% |
| smS_2 | 0.19% | 0.16% | 0.24% | 0.20% | 0.20% | 0.03% |
| sdS | 0.25% | 0.28% | 0.25% | 0.29% | 0.27% | 0.02% |
| aS | 8.04% | 8.18% | 8.24% | 8.70% | 8.29% | 0.29% |
| wS | 0.02% | 0.01% | 0.01% | 0.01% | 0.01% | 0.00% |
| *Total soils* | *18.24%* | *17.71%* | *20.48%* | *19.49%* | *18.98%* | *1.25%* |
| SN | 0.01% | 0.01% | 0.02% | 0.01% | 0.01% | 0.01% |
| WA | 1.28% | 1.28% | 1.25% | 1.27% | 1.27% | 0.02% |
| O | 1.19% | 1.02% | 0.96% | 1.13% | 1.07% | 0.10% |

Table 2-3. Spectral category-specific percentage of occurrences in the SIAM-WELD 2006/ 2007/ 2008/ 2009 test maps at the intermediate level of color quantization, where 48 color names were aggregated into 19 spectral macro-categories by an independent human expert. Adopted acronyms for the 19 spectral macro-categories: refer to Table 2-2.

| | 2006 OAMTRX, Probabilities (%). Rows: SIAM™-WELD 2006, 19 spectral categories; Columns: NLCD 2006, 16 land cover classes. | | | | | | | | | | | | | | | | |
|---|---|---|---|---|---|---|---|---|---|---|---|---|---|---|---|---|---|
| NLCD code | | 11 | 12 | 21 | 22 | 23 | 24 | 31 | 41 | 42 | 43 | 52 | 71 | 81 | 82 | 90 | 95 |
| NLCD class | | OW | PIS | DOS | DLI | DMI | DHI | BL | DF | EF | MF | SS | GH | PH | CC | WW | EHW |
| LCCD-DP1&2 | | B4 | B4 | B3 -> A1 | B3 -> A1 | B3 -> A1 | B3 -> A1 | B3 | A1 | A1 | A1 | A1 -> B3 | A1 -> B3 | A1 | A1 | A2 | A2 |
| sV_HC | | 0.07% | 0.00% | 1.00% | 0.13% | 0.01% | 0.00% | 0.02% | 9.51% | 4.29% | 1.73% | 1.10% | 0.77% | 3.04% | 8.35% | 2.76% | 0.31% | 33.11% |
| aV_HC | | 0.25% | 0.00% | 1.31% | 0.82% | 0.20% | 0.02% | 0.04% | 1.38% | 4.75% | 0.33% | 1.67% | 1.66% | 2.32% | 3.71% | 0.99% | 0.50% | 19.94% |
| wV_HC | | 0.04% | 0.00% | 0.00% | 0.01% | 0.01% | 0.00% | 0.00% | 0.00% | 0.04% | 0.00% | 0.02% | 0.01% | 0.00% | 0.01% | 0.00% | 0.02% | 0.18% |
| sV_MC | | 0.00% | 0.00% | 0.00% | 0.00% | 0.00% | 0.00% | 0.00% | 0.00% | 0.00% | 0.00% | 0.00% | 0.00% | 0.00% | 0.00% | 0.00% | 0.00% | 0.00% |
| aV_MC | | 0.04% | 0.00% | 0.70% | 0.28% | 0.09% | 0.01% | 0.05% | 0.48% | 2.31% | 0.06% | 4.93% | 6.66% | 1.47% | 2.55% | 0.19% | 0.23% | 20.05% |
| sV_LC | | 0.00% | 0.00% | 0.00% | 0.00% | 0.00% | 0.00% | 0.00% | 0.00% | 0.00% | 0.00% | 0.00% | 0.00% | 0.00% | 0.00% | 0.00% | 0.00% | 0.00% |
| aV_LC | | 0.00% | 0.00% | 0.01% | 0.00% | 0.00% | 0.00% | 0.00% | 0.00% | 0.01% | 0.00% | 0.07% | 0.27% | 0.02% | 0.02% | 0.00% | 0.00% | 0.40% |
| wbV_MLC | | 0.01% | 0.00% | 0.07% | 0.07% | 0.09% | 0.03% | 0.04% | 0.00% | 0.31% | 0.00% | 2.54% | 0.79% | 0.02% | 0.13% | 0.01% | 0.01% | 4.12% |
| wdV_MLC | | 0.02% | 0.00% | 0.02% | 0.03% | 0.06% | 0.02% | 0.03% | 0.01% | 0.38% | 0.00% | 0.71% | 0.12% | 0.01% | 0.05% | 0.00% | 0.03% | 1.48% |
| sbS_1 | | 0.01% | 0.00% | 0.09% | 0.04% | 0.04% | 0.02% | 0.57% | 0.00% | 0.01% | 0.00% | 2.88% | 0.90% | 0.03% | 0.38% | 0.01% | 0.01% | 5.00% |
| sbS_2 | | 0.00% | 0.00% | 0.00% | 0.00% | 0.00% | 0.01% | 0.03% | 0.00% | 0.00% | 0.00% | 0.05% | 0.00% | 0.00% | 0.00% | 0.00% | 0.00% | 0.09% |
| smS_1 | | 0.01% | 0.00% | 0.05% | 0.01% | 0.00% | 0.00% | 0.04% | 0.01% | 0.04% | 0.00% | 2.09% | 1.94% | 0.03% | 0.43% | 0.00% | 0.01% | 4.65% |
| smS_2 | | 0.00% | 0.00% | 0.00% | 0.00% | 0.00% | 0.00% | 0.05% | 0.00% | 0.00% | 0.00% | 0.12% | 0.00% | 0.00% | 0.00% | 0.00% | 0.00% | 0.19% |
| sdS | | 0.01% | 0.00% | 0.00% | 0.00% | 0.01% | 0.01% | 0.03% | 0.00% | 0.01% | 0.00% | 0.16% | 0.01% | 0.00% | 0.00% | 0.00% | 0.00% | 0.25% |
| aS | | 0.01% | 0.00% | 0.08% | 0.04% | 0.04% | 0.05% | 0.17% | 0.01% | 0.12% | 0.00% | 5.47% | 1.76% | 0.02% | 0.26% | 0.01% | 0.01% | 8.04% |
| wS | | 0.00% | 0.00% | 0.00% | 0.00% | 0.00% | 0.00% | 0.00% | 0.00% | 0.00% | 0.00% | 0.00% | 0.01% | 0.00% | 0.00% | 0.00% | 0.00% | 0.02% |
| SN | | 0.00% | 0.00% | 0.00% | 0.00% | 0.00% | 0.00% | 0.00% | 0.00% | 0.00% | 0.00% | 0.00% | 0.00% | 0.00% | 0.00% | 0.00% | 0.00% | 0.01% |
| WA | | 1.10% | 0.00% | 0.00% | 0.00% | 0.01% | 0.01% | 0.04% | 0.00% | 0.00% | 0.00% | 0.03% | 0.01% | 0.00% | 0.01% | 0.00% | 0.05% | 1.28% |
| O | | 0.14% | 0.00% | 0.04% | 0.04% | 0.03% | 0.01% | 0.09% | 0.01% | 0.08% | 0.00% | 0.34% | 0.11% | 0.03% | 0.23% | 0.01% | 0.02% | 1.19% |
| | | 1.71% | 0.02% | 3.36% | 1.48% | 0.59% | 0.20% | 1.21% | 11.41% | 12.35% | 2.13% | 22.19% | 15.03% | 6.99% | 16.12% | 3.99% | 1.21% | 100.00% |

Table 2-4. OAMTRX instance generated from a wall-to-wall overlap between the test SIAM-WELD 2006 map of the CONUS with legend A = 19 spectral macro-categories and the reference NLCD 2006 map with legend B = 16 LC class names. Gray entry-pair cells identify the binary relationship R: A $\Rightarrow$ B $\subseteq$ A × B chosen by the independent human expert to guide the interpretation process of the OAMTRX = FrequencyCount(A × B). Statistically independent TQ$^2$Is are CVPAI2(R: A $\Rightarrow$ B) = 0.6769 and OA(OAMTRX) = 96.88%. Adopted acronyms for reference LC classes and test spectral macro-categories are described in Table 2-1 and Table 2-2 respectively.

| | | NLCD Code (class acronym), 16 classes | 41 (DF), 42 (EF), 43 (MF), 52 (SS) <-> B3, 71 (GH) <-> B3, 81 (PH), 82 (CC) | 90 (WV), 95 (EHW) | 21 (DOS) <-> A1, 22 (DLI) <-> A1, 23 (DMI) <-> A1, 24 (DHI) <-> A1, 31 (BL) | 11 (OW), 12 (PIS) | |
|---|---|---|---|---|---|---|---|
| | | LCCS-DP1&2 Code, 4 classes | ≈ A1 | ≈ A2 | ≈ B3 | ≈ B4 | |
| | | LCCS-DP1&2 Class name, 4 classes | Veg terstrl | Veg aqutc | Non-veg terstrl | Non-veg aqutc | Sum per row |
| SIAM™ Intermediate Granularity, 19 Spectral Categories. | | sV_HC | 28.80% | 3.07% | 1.17% | 0.07% | 33.11% |
| | | aV_HC | 15.82% | 1.49% | 2.38% | 0.25% | 19.94% |
| | | wV_HC | 0.08% | 0.03% | 0.03% | 0.04% | 0.18% |
| | | sV_MC | 0.00% | 0.00% | 0.00% | 0.00% | 0.00% |
| | | aV_MC | 18.45% | 0.42% | 1.13% | 0.05% | 20.05% |
| | | sV_LC | 0.00% | 0.00% | 0.00% | 0.00% | 0.00% |
| | | aV_LC | 0.39% | 0.00% | 0.01% | 0.00% | 0.40% |
| | | wbV_MLC | 3.80% | 0.02% | 0.30% | 0.01% | 4.12% |
| | | wdV_MLC | 1.27% | 0.04% | 0.15% | 0.02% | 1.48% |
| | | sbS_1 | 4.21% | 0.01% | 0.76% | 0.01% | 5.00% |
| | | sbS_2 | 0.06% | 0.00% | 0.03% | 0.00% | 0.09% |
| | | smS_1 | 4.53% | 0.01% | 0.10% | 0.01% | 4.65% |
| | | smS_2 | 0.13% | 0.00% | 0.06% | 0.00% | 0.19% |
| | | sdS | 0.18% | 0.00% | 0.06% | 0.01% | 0.25% |
| | | aS | 7.63% | 0.02% | 0.38% | 0.01% | 8.04% |
| | | wS | 0.02% | 0.00% | 0.00% | 0.00% | 0.02% |
| | | SN | 0.00% | 0.00% | 0.00% | 0.00% | 0.01% |
| | | WA | 0.06% | 0.05% | 0.07% | 1.11% | 1.28% |
| | | O | 0.80% | 0.03% | 0.21% | 0.14% | 1.19% |
| | | Sum per column | 86.23% | 5.20% | 6.84% | 1.73% | |

Table 2-5. OAMTRX instance generated from a wall-to-wall overlap between the test SIAM-WELD 2006 map of the CONUS with legend C = 19 spectral macro-categories and the reference NLCD 2006 map with legend B = 2-level 4-class LCCS-DP names. Gray entry-pair cells identify the binary relationship R: C ⇒ B ⊆ C × B chosen by the present authors to guide the interpretation process of the OAMTRX = FrequencyCount(C × B). Statistically independent TQ$^2$Is are CVPAI2(R: C ⇒ B) = 0.7486 and OA(OAMTRX) = 93.09%. Adopted acronyms for reference LC classes and test spectral macro-categories are described in Figure 2-1 and Table 2-2 respectively.

| NLCD class | SIAM1 | p_{SIAM1|NLCD} | SIAM2 | p_{SIAM2|NLCD} | SIAM3 | p_{SIAM3|NLCD} | SIAM4 | p_{SIAM4|NLCD} | SIAM5 | p_{SIAM5|NLCD} |
|---|---|---|---|---|---|---|---|---|---|---|
| Open Water | WA | 0.64 | aV_HC | 0.15 | O | 0.08 | sV_HC | 0.04 | wV_HC | 0.03 |
| Ice/Snow | SN | 0.22 | O | 0.16 | WA | 0.10 | aV_MC | 0.10 | aS | 0.08 |
| Developed, Open | aV_HC | 0.39 | sV_HC | 0.30 | aV_MC | 0.21 | sbS_1 | 0.03 | aS | 0.02 |
| Developed, Low | aV_HC | 0.55 | aV_MC | 0.19 | sV_HC | 0.09 | wbV_MLC | 0.05 | sbS_1 | 0.03 |
| Developed, Medium | aV_HC | 0.33 | wbV_MLC | 0.15 | aV_MC | 0.15 | wdV_MLC | 0.10 | aS | 0.07 |
| Developed, High | aS | 0.24 | wbV_MLC | 0.16 | wdV_MLC | 0.11 | sbS_1 | 0.11 | aV_HC | 0.08 |
| Rock/Sand/Clay | sbS_1 | 0.48 | aS | 0.14 | O | 0.07 | aV_MC | 0.04 | smS_2 | 0.04 |
| Deciduous Forest | sV_HC | 0.83 | aV_HC | 0.12 | aV_MC | 0.04 | O | 0.00 | wdV_MLC | 0.00 |
| Evergreen Forest | aV_HC | 0.38 | sV_HC | 0.35 | aV_MC | 0.19 | wdV_MLC | 0.03 | wbV_MLC | 0.03 |
| Mixed Forest | sV_HC | 0.81 | aV_HC | 0.16 | aV_MC | 0.03 | O | 0.00 | wbV_MLC | 0.00 |
| Shrub/Scrub | aS | 0.25 | aV_MC | 0.22 | sbS_1 | 0.13 | wbV_MLC | 0.11 | smS_1 | 0.09 |
| Grassland/Herbaceous | aV_MC | 0.44 | smS_1 | 0.13 | aS | 0.12 | aV_HC | 0.11 | sbS_1 | 0.06 |
| Pasture/Hay | sV_HC | 0.43 | aV_HC | 0.33 | aV_MC | 0.21 | sbS_1 | 0.00 | smS_1 | 0.00 |
| Cultivated Crops | sV_HC | 0.52 | aV_HC | 0.23 | aV_MC | 0.16 | smS_1 | 0.03 | sbS_1 | 0.02 |
| Woody Wetlands | sV_HC | 0.69 | aV_HC | 0.25 | aV_MC | 0.05 | wbV_MLC | 0.00 | O | 0.00 |
| Herbaceous Wetlands | aV_HC | 0.41 | sV_HC | 0.26 | aV_MC | 0.19 | WA | 0.04 | wdV_MLC | 0.03 |

Table 2-6. Class-conditional probability p(SIAM-WELD_t | NLCD_r), t = 1, ..., TC = |A| = 19 color names, r = 1, ... RC = |B| = 16 LC class names. For each NLCD 2006 reference map's LC class, the five best-matching SIAM-WELD 2006 test map's spectral macro-categories, belonging to the finite set A of 19 spectral macro-categories, are shown as SIAM1 to SIAM5.

| SIAM | NLCD 1 | p_{NLCD1|SIAM} | NLCD 2 | p_{NLCD2|SIAM} | NLCD 3 | p_{NLCD3|SIAM} | NLCD 4 | p_{NLCD4|SIAM} | NLCD 5 | p_{NLCD5|SIAM} |
|---|---|---|---|---|---|---|---|---|---|---|
| sV_HC | Deciduous Forest | 0.29 | Cultivated Crops | 0.25 | Evergreen Forest | 0.13 | Pasture/Hay | 0.09 | Woody Wetlands | 0.08 |
| aV_HC | Evergreen Forest | 0.24 | Cultivated Crops | 0.19 | Pasture/Hay | 0.12 | Shrub/Scrub | 0.08 | Grassland/Herbace | 0.08 |
| wV_HC | Open Water | 0.25 | Evergreen Forest | 0.20 | Herbaceous Wetlands | 0.12 | Shrub/Scrub | 0.12 | Developed, Mediur | 0.08 |
| sV_MC | Deciduous Forest | 0.33 | Cultivated Crops | 0.17 | Pasture/Hay | 0.15 | Grassland/Herbaceo | 0.08 | Woody Wetlands | 0.06 |
| aV_MC | Grassland/Herbaceo | 0.33 | Shrub/Scrub | 0.25 | Cultivated Crops | 0.13 | Evergreen Forest | 0.12 | Pasture/Hay | 0.07 |
| sV_LC | Shrub/Scrub | 0.04 | Evergreen Forest | 0.03 | Grassland/Herbaceous | 0.01 | Woody Wetlands | 0.01 | Cultivated Crops | 0.01 |
| aV_LC | Grassland/Herbaceo | 0.68 | Shrub/Scrub | 0.18 | Cultivated Crops | 0.05 | Pasture/Hay | 0.04 | Evergreen Forest | 0.02 |
| wbV_MLC | Shrub/Scrub | 0.62 | Grassland/Herbaceous | 0.19 | Evergreen Forest | 0.08 | Cultivated Crops | 0.03 | Developed, Mediur | 0.02 |
| wdV_MLC | Shrub/Scrub | 0.48 | Evergreen Forest | 0.26 | Grassland/Herbaceous | 0.08 | Developed, Medium | 0.04 | Cultivated Crops | 0.03 |
| sbS_1 | Shrub/Scrub | 0.58 | Grassland/Herbaceous | 0.18 | Rock/Sand/Clay | 0.11 | Cultivated Crops | 0.08 | Developed, Open | 0.02 |
| sbS_2 | Shrub/Scrub | 0.58 | Rock/Sand/Clay | 0.27 | Developed, High | 0.06 | Developed, Medium | 0.03 | Grassland/Herbace | 0.02 |
| smS_1 | Shrub/Scrub | 0.45 | Grassland/Herbaceous | 0.42 | Cultivated Crops | 0.09 | Developed, Open | 0.01 | Rock/Sand/Clay | 0.01 |
| smS_2 | Shrub/Scrub | 0.66 | Rock/Sand/Clay | 0.25 | Grassland/Herbaceous | 0.03 | Developed, Low | 0.01 | Developed, Mediur | 0.01 |
| sdS | Shrub/Scrub | 0.63 | Rock/Sand/Clay | 0.13 | Developed, High | 0.05 | Evergreen Forest | 0.04 | Grassland/Herbace | 0.04 |
| aS | Shrub/Scrub | 0.68 | Grassland/Herbaceous | 0.22 | Cultivated Crops | 0.03 | Rock/Sand/Clay | 0.02 | Evergreen Forest | 0.01 |
| wS | Grassland/Herbaceo | 0.71 | Shrub/Scrub | 0.14 | Cultivated Crops | 0.06 | Pasture/Hay | 0.03 | Evergreen Forest | 0.02 |
| SN | Ice/Snow | 0.47 | Rock/Sand/Clay | 0.35 | Grassland/Herbaceous | 0.06 | Shrub/Scrub | 0.03 | Cultivated Crops | 0.03 |
| WA | Open Water | 0.86 | Herbaceous Wetlands | 0.04 | Rock/Sand/Clay | 0.03 | Shrub/Scrub | 0.03 | Grassland/Herbace | 0.01 |
| O | Shrub/Scrub | 0.28 | Cultivated Crops | 0.20 | Open Water | 0.12 | Grassland/Herbaceo | 0.09 | Rock/Sand/Clay | 0.07 |

Table 2-7. Class-conditional probability p(NLCD_r | SIAM-WELD_t), t = 1, ..., TC = |A| = 19 color names, r = 1, ... RC = |B| = 16 LC class names. For each SIAM-WELD 2006 test map's spectral macro-category, the five best-matching NLCD 2006 reference map's LC classes, belonging to the finite set B of 16 LC classes, are shown as NLCD1 to NLCD5.

| | | NLCD code | 11 | 12 | 21 | 22 | 23 | 24 | 31 | 41 | 42 | 43 | 52 | 71 | 81 | 82 | 90 | 95 | |
|---|---|---|---|---|---|---|---|---|---|---|---|---|---|---|---|---|---|---|---|
| | Wyoming Basin Ecoregion, 2006 OAMTRX. Probabilities (%). Rows: SIAM™-WELD 2006, 19 spectral categories; Columns: NLCD 2006, 16 land cover classes. | | | | | | | | | | | | | | | | | | |
| | | NLCD class | OW | PIS | DOS | DLI | DMI | DHI | BL | DF | EF | MF | SS | GH | PH | CC | WW | EHW | |
| | | LCCD-DP1&2 | B4 | B4 | B3 -> A1 | B3 -> A1 | B3 -> A1 | B3 -> A1 | B3 | A1 | A1 | A1 | A1 -> B3 | A1 -> B3 | A1 | A1 | A2 | A2 | |
| SIAM™ Intermediate Granularity, 19 Spectral Categories. | | sV_HC | 0.00% | 0.00% | 0.03% | 0.01% | 0.00% | 0.00% | 0.00% | 0.03% | 0.02% | 0.01% | 0.06% | 0.02% | 1.04% | 0.23% | 0.09% | 0.11% | 1.65% |
| | | aV_HC | 0.02% | 0.00% | 0.05% | 0.03% | 0.01% | 0.00% | 0.00% | 0.07% | 0.43% | 0.02% | 0.41% | 0.13% | 1.05% | 0.16% | 0.38% | 0.35% | 3.11% |
| | | wV_HC | 0.01% | 0.00% | 0.00% | 0.00% | 0.00% | 0.00% | 0.00% | 0.00% | 0.01% | 0.00% | 0.01% | 0.00% | 0.00% | 0.00% | 0.01% | 0.00% | 0.05% |
| | | sV_MC | 0.00% | 0.00% | 0.00% | 0.00% | 0.00% | 0.00% | 0.00% | 0.00% | 0.00% | 0.00% | 0.00% | 0.00% | 0.00% | 0.00% | 0.00% | 0.00% | 0.00% |
| | | aV_MC | 0.01% | 0.00% | 0.06% | 0.02% | 0.00% | 0.00% | 0.00% | 0.08% | 0.68% | 0.01% | 3.47% | 1.21% | 0.60% | 0.06% | 0.23% | 0.42% | 6.85% |
| | | sV_LC | 0.00% | 0.00% | 0.00% | 0.00% | 0.00% | 0.00% | 0.00% | 0.00% | 0.00% | 0.00% | 0.00% | 0.00% | 0.00% | 0.00% | 0.00% | 0.00% | 0.00% |
| | | aV_LC | 0.00% | 0.00% | 0.00% | 0.00% | 0.00% | 0.00% | 0.00% | 0.00% | 0.00% | 0.00% | 0.06% | 0.01% | 0.00% | 0.00% | 0.00% | 0.00% | 0.07% |
| | | wbV_MLC | 0.01% | 0.00% | 0.04% | 0.03% | 0.01% | 0.00% | 0.00% | 0.00% | 0.18% | 0.00% | 2.35% | 0.55% | 0.04% | 0.02% | 0.05% | 0.07% | 3.35% |
| | | wdV_MLC | 0.01% | 0.00% | 0.01% | 0.00% | 0.00% | 0.00% | 0.00% | 0.00% | 0.14% | 0.00% | 0.57% | 0.05% | 0.01% | 0.00% | 0.01% | 0.01% | 0.81% |
| | | sbS_1 | 0.01% | 0.00% | 0.12% | 0.04% | 0.01% | 0.00% | 0.69% | 0.00% | 0.01% | 0.00% | 16.90% | 5.66% | 0.09% | 0.04% | 0.04% | 0.11% | 23.72% |
| | | sbS_2 | 0.00% | 0.00% | 0.00% | 0.00% | 0.00% | 0.00% | 0.00% | 0.00% | 0.00% | 0.00% | 0.02% | 0.01% | 0.00% | 0.00% | 0.00% | 0.00% | 0.03% |
| | | smS_1 | 0.00% | 0.00% | 0.12% | 0.01% | 0.00% | 0.00% | 0.03% | 0.00% | 0.06% | 0.00% | 32.71% | 4.11% | 0.04% | 0.01% | 0.02% | 0.09% | 37.20% |
| | | smS_2 | 0.00% | 0.00% | 0.00% | 0.00% | 0.00% | 0.00% | 0.01% | 0.00% | 0.00% | 0.00% | 0.04% | 0.02% | 0.00% | 0.00% | 0.00% | 0.00% | 0.07% |
| | | sdS | 0.01% | 0.00% | 0.00% | 0.00% | 0.00% | 0.00% | 0.00% | 0.00% | 0.00% | 0.00% | 0.08% | 0.01% | 0.00% | 0.00% | 0.00% | 0.00% | 0.10% |
| | | aS | 0.02% | 0.00% | 0.17% | 0.07% | 0.01% | 0.00% | 0.08% | 0.00% | 0.13% | 0.00% | 17.86% | 3.25% | 0.05% | 0.02% | 0.03% | 0.10% | 21.79% |
| | | wS | 0.00% | 0.00% | 0.00% | 0.00% | 0.00% | 0.00% | 0.00% | 0.00% | 0.00% | 0.00% | 0.00% | 0.00% | 0.00% | 0.00% | 0.00% | 0.00% | 0.00% |
| | | SN | 0.00% | 0.00% | 0.00% | 0.00% | 0.00% | 0.00% | 0.00% | 0.00% | 0.00% | 0.00% | 0.00% | 0.00% | 0.00% | 0.00% | 0.00% | 0.00% | 0.00% |
| | | WA | 0.54% | 0.00% | 0.00% | 0.00% | 0.00% | 0.00% | 0.01% | 0.00% | 0.00% | 0.00% | 0.06% | 0.01% | 0.00% | 0.00% | 0.01% | 0.01% | 0.64% |
| | | O | 0.02% | 0.00% | 0.01% | 0.01% | 0.00% | 0.00% | 0.01% | 0.00% | 0.01% | 0.00% | 0.31% | 0.06% | 0.05% | 0.03% | 0.02% | 0.03% | 0.56% |
| | | | 0.65% | 0.00% | 0.61% | 0.22% | 0.04% | 0.01% | 0.83% | 0.18% | 1.67% | 0.04% | 74.91% | 15.10% | 2.97% | 0.57% | 0.89% | 1.30% | 100.00% |

Table 2-8. OAMTRX instance generated from a wall-to-wall overlap over the Wyoming Basin Ecoregion between the test SIAM-WELD 2006 map with legend A = 19 spectral macro-categories and the reference NLCD 2006 map with legend B = 16 LC class names. Gray squares identify the binary relationship R: A ⇒ B ⊆ A × B chosen by the independent human expert to guide the interpretation process of the OAMTRX = FrequencyCount(A × B), same as in Table 2-4. The Wyoming Basin Ecoregion is predominantly desertic. It is classified as LC class "*Scrub/Shrub*" (SS) or LC class "*Grassland/Herbaceous*" (GH) in the NLCD 2006 reference map (refer to Table 2-1), and predominantly as spectral macro-categories of bare soil (sbS_1, SmS_1, aS) in the SIAM-WELD 2006 test map (refer to Table 2-2). This phenomenon of large-scale "conceptual mismatch" between the NLCD 2006 and SIAM-WELD 2006 thematic maps is discussed thoroughly in Section 4.3.